\pdfoutput=1

\documentclass[11pt]{article}
\usepackage{tcolorbox}
\usepackage{listings}
\usepackage{xspace}
\usepackage{soul}
\usepackage{colortbl}
\usepackage{xcolor}
\usepackage{arydshln}
\usepackage{tabularx}
\usepackage{caption}
\usepackage{tcolorbox}

\tcbuselibrary{skins, breakable, theorems}
\usepackage{colortbl}
\definecolor{darkgreen}{rgb}{0,0.5,0} 
\definecolor{purple}{rgb}{1,0,1} 
\definecolor{todocolor}{rgb}{0.9,0.1,0.1} 
\definecolor{fixcolor}{rgb}{0.1,0.7,0.3} 
\definecolor{wycolor}{rgb}{0.9,0.1,0.1} 
\definecolor{hycolor}{rgb}{0.7,0.7,0.3} 
\definecolor{zwcolor}{rgb}{1,0,1} 
\definecolor{gycolor}{rgb}{0.06,0.93,0.93} 
\newcount\DraftStatus  
\newcommand{\nbc}[3]{\ifnum\DraftStatus=1
	{\colorbox{#3}{\bfseries\sffamily\scriptsize\textcolor{white}{#1}}}
	{\textcolor{#3}{\sf\small$\blacktriangleright$\emph{#2}$\blacktriangleleft$}}
\fi}

\newtcolorbox{promptbox}[1][]{%
  colback=gray!5!white, colframe=gray!75!black, sharp corners, 
  boxrule=0.5mm, top=10pt, bottom=10pt, left=10pt, right=10pt, breakable, title={#1}
}

\newcommand{\draftnote}[2]{\ifnum\DraftStatus=1
	\marginpar{
		\tiny\raggedright
		\hbadness=10000
		\def\baselinestretch{0.8}
		\textcolor{#1}{\textsf{\hspace{0pt}#2}}}
\fi}

\usepackage[final]{acl}

\usepackage{graphicx}

\usepackage{times}
\usepackage{latexsym}
\usepackage{amsmath}
\usepackage{amssymb}
\usepackage{mathtools}
\usepackage{amsthm}
\usepackage{enumitem}
\usepackage{multirow}
\usepackage{makecell}

\usepackage{booktabs}
\usepackage{arydshln}
\usepackage{algorithm}
\usepackage{algpseudocode}
\usepackage{pifont}
\usepackage{colortbl}
\usepackage{soul}
\usepackage[T1]{fontenc}
\usepackage[utf8]{inputenc}
\usepackage{microtype}
\usepackage{inconsolata}
\usepackage{subcaption}
\usepackage{titlesec}
\usepackage{lipsum}
\usepackage{setspace}

\newcommand\blfootnote[1]{%
  \begingroup
  \renewcommand\thefootnote{}\footnote{#1}%
  \addtocounter{footnote}{-1}%
  \endgroup
}

\newcommand{\method}{\textsc{CrowdSelect}\xspace}

\title{{\includegraphics[height=1.5em]{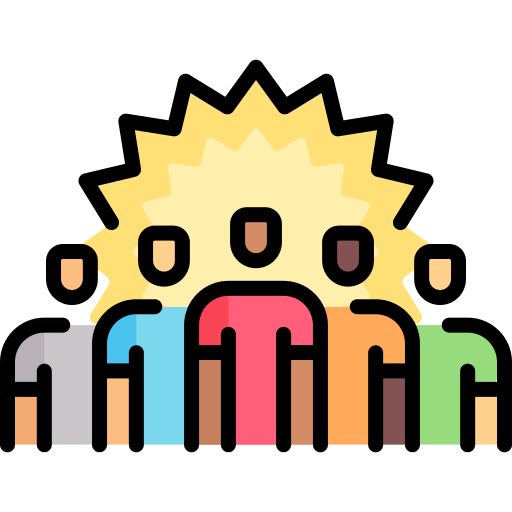}}~\method: Synthetic Instruction Data Selection\\with Multi-LLM Wisdom}

\author{
 \textbf{Yisen Li \textsuperscript{1\textdagger}},
 \textbf{Lingfeng Yang\textsuperscript{1\textdagger}},
 \textbf{Wenxuan Shen\textsuperscript{2}},
 \textbf{Pan Zhou\textsuperscript{1*}},
 \textbf{Yao Wan\textsuperscript{1*}},\\
 \textbf{Weiwei Lin\textsuperscript{2}},
 \textbf{Dongping Chen\textsuperscript{1}\textsuperscript{$\ddagger$}}\\
 \textsuperscript{1}
 Huazhong University of Science and Technology\\
 \textsuperscript{2}
 South China University of Technology\\
 \texttt{\{panzhou,wanyao\}@hust.edu.cn}, \texttt{linww@scut.edu.cn}
 }

\begin{document}                                
\maketitle

\blfootnote{\textdagger \ Equal Contribution. \textsuperscript{$\ddagger$} Project Leader.}
\blfootnote{* Corresponding Authors.}

\begin{abstract}
\setstretch{0.87}
Distilling advanced Large Language Models' instruction-following capabilities into smaller models using a selected subset has become a mainstream approach in model training. While existing synthetic instruction data selection strategies rely mainly on single-dimensional signals (\emph{i.e.}, reward scores, model perplexity), they fail to capture the complexity of instruction-following across diverse fields. Therefore, we investigate more diverse signals to capture comprehensive instruction-response pair characteristics and propose three foundational metrics that leverage Multi-LLM wisdom, informed by \textit{(1)} diverse LLM responses and \textit{(2)} reward model assessment. Building upon base metrics, we propose \method, an integrated metric incorporating a clustering-based approach to maintain response diversity.
Our comprehensive experiments demonstrate that our foundation metrics consistently improve performance across 4 base models on MT-bench and Arena-Hard. \method, efficiently incorporating all metrics, achieves \emph{state-of-the-art} performance in both Full and LoRA fine-tuning, showing improvements of 4.81\% on Arena-Hard and 11.1\% on MT-bench with Llama-3.2-3b-instruct. We hope our findings will bring valuable insights for future research in this direction. Code are available at \href{https://github.com/listentm/crowdselect}{https://github.com/listentm/crowdselect}.
\end{abstract}
\section{Introduction}
In recent years, Large Language Models (LLMs) \citep{achiam2023gpt,jaech2024openai,team2024gemma, guo2025deepseek} have demonstrated remarkable capability in following user instructions to generate coherent and contextually helpful responses \citep{jiang2023followbench, zheng2023judging, wen2024benchmarking}. Yet, the computational overhead for instruction tuning and massive parameter sizes of these models create a considerable barrier to practical deployment \citep{Peng2023InstructionTW}. To address this, many approaches distill the instruction-following ability of advanced LLMs into smaller, more efficient models through a small-scale instruction tuning process with synthetic responses \citep{Xia2024LESSSI, zhou2024lima}. 

A critical bottleneck, however, lies in selecting the optimal data for this distillation process. Most existing data selection methods rely on predefined rules \citep{chen2023alpagasus}, automated single-dimensional signals — such as reward scores \citep{wu2024rose,lambert2024rewardbench} or difficulty metrics \citep{li2023quantity,Li2024SuperfilteringWD} — to identify valuable examples for fine-tuning. While effective to some extent, such narrow signals may overlook essential nuances of user instructions, especially when instructions contain challenges from diverse fields \citep{Hndler2023BalancingAA,Feng2025WhenOL}. This raises a fundamental question: \textit{``Can we leverage multi-dimensional signals to better reflect the various facets of each sample for more effective instruction tuning data selection?''}

\begin{figure}[!t]
    \centering
    \includegraphics[width=1\linewidth]{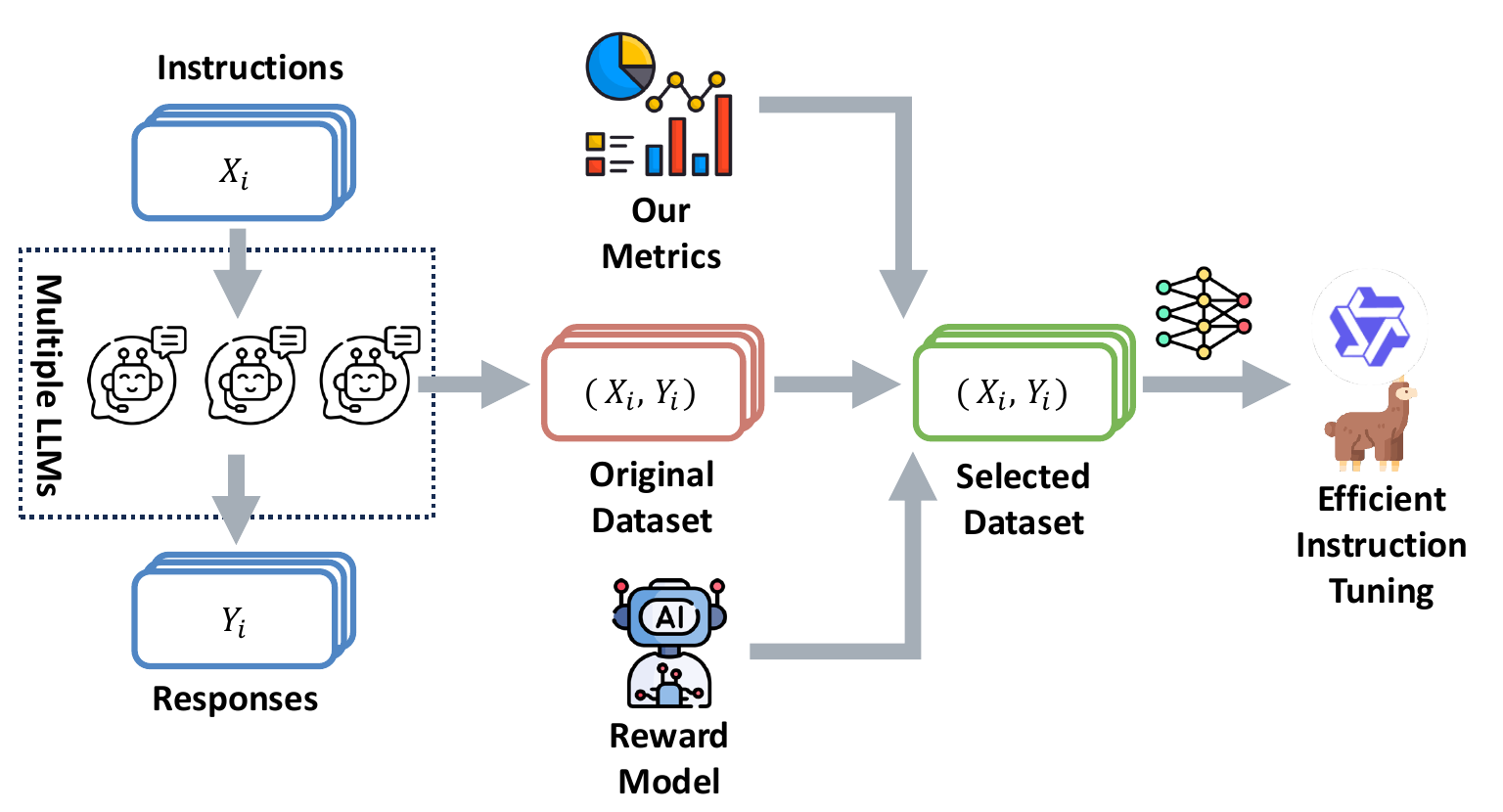}
    \caption{A demonstration of instruction tuning with selected synthetic instruction-response pairs.}
    \vspace{-1em}
    \label{fig:problem-formulation}
\end{figure}

Inspired by previous works that leverage Multi-LLM collaboration \citep{guo2024embodied, lu2024merge}, we take an explorative step towards more robust and comprehensive data selection by introducing \method, a framework that treats pre-collected Multiple LLMs' responses and their reward scores as different reflections of the instruction to leverage Multi-LLM Wisdom. Instead of treating each instruction–response pair in isolation — typically with just a single model — our method aggregates multiple responses for each instruction from a diverse set of LLMs. Crucially, we also factor in each response’s score provided by various reward models. This multi-view setup captures more \textit{``facets''} of each instruction, illuminating subtle differences in how various models handle the same query. Based on these observations, we propose three base explorative metrics:
\begin{itemize}[leftmargin=*, itemsep=0pt]
    \item \textbf{Difficulty -} Identifies instructions on which the majority of models struggle, surfacing challenging prompts critical to learning.
    \item \textbf{Separability –} Highlights instructions whose response quality exhibits high variance across models, making them especially useful for differentiating stronger from weaker capabilities.
    \item \textbf{Stability –} Measures how consistently model performance follows expected size-based ranking across families, ensuring the selected data helps reinforce well-grounded alignment signals.
\end{itemize}
Our exploratory experiments in full fine-tuning (FFT) and low-rank adaptation (LoRA) \citep{hu2021lora} experiments on LLama-3.2-3b-base/instruct \citep{dubey2024llama} and Qwen-2.5-3b-base/instruct \citep{yang2024qwen2} demonstrate the robustness and efficacy of our proposed metrics through significant performance gaps between \textit{top-scored} and \textit{bottom-scored} data subset fine-tuning, with potential further improvements through metric combination. 

Subsequently, we propose \method that combines these metrics with a clustering strategy to preserve diversity and explore the upper bound of leveraging Multi-LLM wisdom to identify a compact yet high-impact subset of instruction-response data. Experimental results show that models fine-tuned on our selected subset significantly outperform baselines and previous \emph{state-of-the-art} data selection methods, achieving improvements of 4.81\% on Arena-Hard and 11.1\% on MT-bench with Llama-3.2-3b-instruct. Furthermore, \method achieves \emph{state-of-the-art} performance across four models on two benchmarks, demonstrating both the generalizability and robustness of our selected data and methodology, paving a new dimension for efficient instruction tuning.

Our contributions are summarized as follows:
\begin{itemize}[leftmargin=*, itemsep=0pt]
    \item \textbf{Investigation of Multi-LLM Wisdom in Instruction Data Selection.} We propose a novel approach that utilizes multiple synthesized responses from different LLMs for each instruction, enhancing the diversity and quality of data.
    \item \textbf{Novel Metrics and Methods.} We design three new explorative base metrics—\textit{Difficulty}, \textit{Separability}, and \textit{Stability}—that leverage multi-LLM responses and reward scores as more comprehensive signals, and combine them into \method to explore the upper bound in selecting high-quality data for instruction tuning.
    \item \textbf{\emph{State-of-the-art} Performance.} We demonstrate that combining our metrics and clustering techniques for data selection leads to a new SOTA in efficient instruction tuning in both Llama-3.2-3b and Qwen-2.5-3b.
\end{itemize}

\section{Related Work}
\label{related_work}

\paragraph{Instruction Tuning Data Selection.}
Instruction Tuning stands out to be a method to solve the gap between pre-trained knowledge and real-world user scenarios \citep{ouyang2022training,bai2022training}. Recent efforts like Vicuna \citep{Peng2023InstructionTW} and LIMA \citep{zhou2024lima} demonstrate high performance with a carefully selected small dataset, highlighting the growing importance of efficient instruction tuning.
Three key metrics determine instruction data quality: \textit{Difficulty}, \textit{Quality}, and \textit{Diversity}. \textit{Difficulty}, focusing mainly on the question side, is considered more valuable for model learning \citep{li2023quantity,Li2024SuperfilteringWD,liu2024selectit,lee2024instruction,wang2024survey}. \textit{Quality}, mainly addressing the response side, measures the helpfulness and safety of model responses, typically assessed using LLM evaluators \citep{chen2023alpagasus,chen2024mllm,Liu2024SelectITSI,ye2024justice}, reward models \citep{son2024llm,lambert2024rewardbench}, and gradient similarity search \citep{Xia2024LESSSI}. \textit{Diversity} also plays a crucial role in covering various instruction formats and world knowledge, primarily improving model robustness \citep{bukharin2023data, wang2024diversity}.

\paragraph{Data Synthesis for Instruction Tuning.}
While the development of LLMs initially relied on human-curated instruction datasets for instruction tuning \citep{zheng2023lmsys,Zhao2024WildChat1C,Lightman2023LetsVS}, this approach proved time-consuming and labor-intensive, particularly as the complexity and scope of target tasks increased \citep{Demrozi2023ACR,Wang2021WantTR}. Consequently, researchers began exploring the use of frontier LLMs to generate synthetic instruction datasets, aiming to both address these scalability challenges  \citep{Ding2023EnhancingCL,Chen2023ShareGPT4VIL,Chen2024ShareGPT4VideoIV} and leverage  models' advanced capabilities in developing next-generation foundation models \citep{Burns2023WeaktoStrongGE,Charikar2024QuantifyingTG}.
Recent advancements streamline this process by utilizing instructions directly from pre-trained LLMs with simple prompt templates \citep{Xu2024MagpieAD,Chen2024GenQAGM,Zhang2024ProVisionPS}, significantly reducing the required custom design from human effort.

\paragraph{Deriving Crowded Wisdom from Multi-LLM.} Single LLM's response to a question face limitations in its representation of data (particularly cutting-edge knowledge) \citep{lazaridou2021mind,dhingra2022time,kasai2023realtime}, skills (as no single LLM is universally optimal \textit{empirically}) \citep{sun2022paradigm,liang2022holistic,chen2024interleaved}, and diverse perspectives \citep{Feng2025WhenOL}. Previous work has demonstrated that \textit{online} multi-LLM wisdom (also known as compositional agent frameworks \citep{gupta2023visual}) tends to outperform single models across various domains, providing more comprehensive and reflective solution on complex downstream tasks \citep{wang2024mixture,wu2023autogen,li2023camel,ouyang2025nvagent,gui2025uicopilot}. \textit{Offline} crowded wisdom, where data are pre-collected rather than real-time inference, also show potential in model alignment \citep{gallego2024refined,rafailov2023direct,meng2025simpo} and benchmark construction \citep{ni2024mixeval,Ni2024MixEvalXAE}. In this paper, we pioneer the use of \textit{offline} multi-LLM wisdom for instruction data selection by utilizing these LLMs' responses and their reward score as \textit{reflections} to measure instruction-response pairs' \textit{Difficulty} and \textit{Quality}.

\begin{figure*}[!t]
    \centering
    \includegraphics[width=1\linewidth]{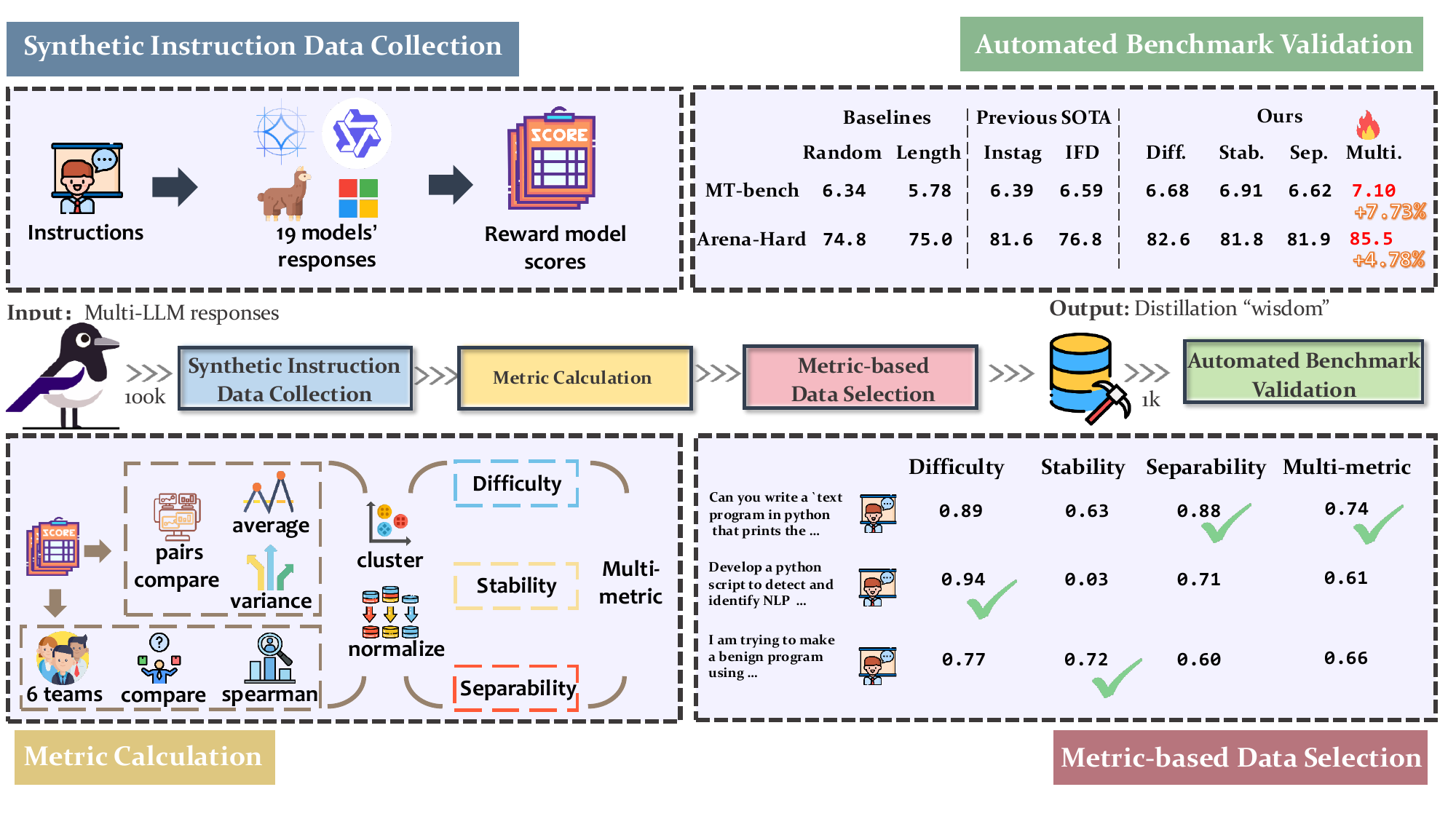}
    \vspace{-2.5em}
    \caption{The overall pipeline of our \method, which innovatively leverages metrics calculated from multiple facets of instructions using pre-collected synthesized responses from various LLMs and their corresponding reward model scores. We enhance data selection through clustering for diversity and metric combination to explore the method's potential. Finally, we evaluate the effectiveness of our selected instruction subset through FFT or LoRA fine-tuning \citep{hu2021lora} for efficient instruction tuning.}
    \vspace{-1em}
    \label{fig:problem-definition}
\end{figure*}

\section{Methodology}
\label{method}
We begin by defining our synthetic data selection task and proposing three foundational metrics that utilize responses and assessment scores from multiple advanced LLMs. Building on these metrics, we introduce \method, which employs diversity-preserving clustering to investigate the upper limits of Multi-LLM Wisdom. An overview of our pipeline is provided in Figure \ref{fig:problem-definition}.

\subsection{Preliminaries}
We formulate the instruction quality as the consensus among $N$ LLMs. Given an instruction-tuning dataset, we extract all instructions from the dataset to form instruction dataset $Q$. For each instruction $q_i \in Q$, a response set $R_i$ is obtained by querying multiple LLMs. 
An assessment model then evaluates the responses in $R_i$ to produce a score set $C^M_i$ according to metrics $M$. For simplicity, the index $M$ will be omitted unless otherwise noted. We define the top-$k$ instruction subset for metric $M$ as follows:
\begin{equation}\label{eq1}
S^M_k = \underset{S\subset\mathcal{S},|S|=k}{\operatorname*{\arg\max}}M(C^M_i)\,,
\end{equation}
where $S^M_k$ consists of the $k$ instructions that maximize the metric $M$.

The corresponding response $r^M_i$ for each instruction $q^M_i$ from the instruction subset $S^M_k$ is subsequently obtained by
\begin{equation}\label{eq2}
r_i^M = \mathrm{Top}(R_i,C_i^M)\,,
\end{equation}
where $\mathrm{Top}(R_i^S,C_i^M)$ denotes the best responses in $r^S_i$ ranked by $C^M_i$.
The produced instruction-answer subset $\hat{Q} = \{(r^M_i, q^M_i)\}$ is then utilized for fine-tuning as an alternative of the original dataset.

\subsection{Base Metrics}
We introduce three new base metrics that incorporate multiple LLM responses and their corresponding reward scores as distinct \textit{``facets''} to assess the value of each sample.

\paragraph{Difficulty.}
The difficulty score $C^{dif}$ is defined as the negative mean of all model response scores for a given instruction, calculated as follows:
\begin{equation}\label{eq3}
C^{dif} = - \frac{\sum C^M_i}{N}\,.
\end{equation}
Higher \textit{difficulty} indicates more challenging instructions. This metric is particularly well-suited for fine-tuning on reasoning tasks, \emph{e.g.} mathematics and planning, where the goal is often to improve performance on complex problems. By focusing on instructions with higher \textit{difficulty}, we prioritize examples that are likely to be answered incorrectly by the majority of models. This ensures that the fine-tuning dataset includes a substantial proportion of challenging instructions, maximizing the model's exposure to difficult material and potentially leading to greater improvements in performance.

\paragraph{Separability.}
The separability score $C^{sep}$ is defined as the score variance, which is the variance of all the response scores for an instruction:
\begin{equation}\label{eq4}
C^{sep} = \operatorname{var}(C^M_i)\,.
\end{equation}
Higher \textit{Separability} indicates that a considerable proportion of models cannot perform well on the instruction, thus this instruction is more effective in differentiating between models. This characteristic makes the \textit{Separability} particularly well-suited for curating datasets of knowledge remembering or preference alignment. In such datasets, some models may exhibit strong performance while others struggle. By selecting instructions with high separability, we prioritize examples that effectively distinguish between these varying levels of competence. These \textit{``discriminatory''} examples are valuable because they provide the fine-tuned model with opportunities to learn from the specific challenges that differentiate successful models from less successful ones. Focusing on these examples enforces the fine-tuned model to handle the nuances and complexities that separate high-performing models.

\paragraph{Stability.}
\textit{Stability} is defined as the average spearman factor, which is the mean of five spearman factors, corresponding to five model families. The spearman factor is calculated based on $r^a$ and $r^b$:
\begin{equation}\label{eq5}
    \scriptstyle
    \frac{\frac{1}{n}\sum_{i=1}^n\left(r_i^a-\overline{r^a}\right)\cdot\left(r_i^b-\overline{r^b}\right)}{\sqrt{\left(\frac{1}{n}\sum_{i=1}^n\left(r_i^a-\overline{r^a}\right)^2\right)\cdot\left(\frac{1}{n}\sum_{i=1}^n\left(r_i^b2-\overline{r^b}\right)^2\right)}}\,.
\end{equation}
\begin{itemize}
  \item $r^a$ refers to the original ranking within a model family, where models with larger parameters are theoretically ranked higher, naturally aligning with the performance rank.
  \item $r^b$ is determined by the rank of models based on their response quality (\emph{e.g.}, if LLaMA-3B has a response score of 9 and LLaMA-8B has a response score of 7, then 3B ranks higher than 8B within the LLaMA family).
\end{itemize}
\textit{Stability} effectively captures how well performance rankings align with expected model size rankings using Spearman's rank correlation \citep{schober2018correlation}, making it robust to variations in score scales and non-linear relationships. Averaging across model families further strengthens the robustness of the score, alleviating performance gaps among model families.

\subsection{\method: Explore the Upperbound with Multi-LLM Wisdom}
\paragraph{Diversity Preservation with Clustering.} To facilitate clustering, all instructions were embedded into a fixed-dimensional latent space using a pre-trained embedding model. Within each cluster, instructions were then ranked with the given metric, and the highest-ranked instructions were selected. To avoid over-representing dominant clusters and neglecting potentially valuable information contained within smaller or less frequent clusters, we draw equally from each cluster to form a more robust and generalizable subset.

\paragraph{Multi-metric Integration.}
Accompanying with the cluster-based selection strategy, we also introduce a multi-metric approach to leverage the diverse information captured by our three foundation metrics. Each instruction-response pair is thus characterized by a vector of associated scores, reflecting its various attributes. However, these metrics exhibit different distributions, ranges, and magnitudes. Therefore, we employ a three-stage normalization process to ensure equitable contribution from each metric.

Specifically, each metric score is standardized to standard normal distribution. The standardized scores are then normalized to $[0, 1]$ using a min-max scaling approach. Finally, to further refine the distribution and mitigate the impact of potential outliers, we apply a quantile transformation that maps the normalized scores to a uniform distribution between $[0, 1]$.
\begin{equation}\label{eq6}
    Z^M_i=\frac{(C^M_i-\mu^M)}{\sigma^M}\,,
\end{equation}
\begin{equation}\label{eq7}
    N^M_i = \frac{(Z^M_i - min(Z^M))}{(max(Z^M) - min(Z^M))}\,,
\end{equation}
\begin{equation}\label{eq8}
    \rho^M_i = \operatorname{quant}(N^M_i|N^M)\,.
\end{equation}
Following this normalization procedure, we aggregate the transformed scores into a single multi-metric score $\hat{C}$ for each instruction-response pair. This aggregation is performed using a weighted sum of the proposed metrics:
\begin{equation}\label{eq9}
    \hat{C}_i = \sum_j w_i * \rho^{M_j}_i\,,
\end{equation}
where $\scriptstyle \rho^{M_j}_i$ represents the quantile-transformed scores for metric $j$, and $w_i$ is the corresponding weight assigned to each metric. This weighted multi-metric approach, combined with the preceding normalization steps, ensures a balanced and robust data selection process that leverages the complementary information provided by all metrics.

\section{Experiment}
\label{experiment}
We begin by validating our base metrics through comparative experiments on the top- and bottom-scored data subsets. Next, we evaluate \method against existing baselines and \emph{state-of-the-art} approaches. Finally, we perform an ablation study to assess the contributions of each sub-module within \method.

\begin{table*}[!t]
\centering
\setlength{\tabcolsep}{6pt}
\caption{Validation of our three foundation metrics on full fine-tuning Llama-3.2-3b-base with \textit{top-scored ($\uparrow$)} and \textit{bottom-scored ($\downarrow$)} instruction selection and different response selection strategy. Best and second results for each metric are in \textbf{bold} and \underline{underline}.}
\vspace{-1em}
\resizebox{\textwidth}{!}{
\begin{tabular}{ccccccccc}
\toprule[1.5pt]
\multirow{2}{*}{\textbf{Strategy}} & \multirow{2}{*}{\textbf{DirectScore}} & \multicolumn{2}{c}{\textbf{Difficulty}} & \multicolumn{2}{c}{\textbf{Separability}} & \multicolumn{2}{c}{\textbf{Stability}} & \multirow{2}{*}{\textbf{Multi}} \\
& & \textbf{$\downarrow$} & \textbf{$\uparrow$} & \textbf{$\downarrow$} & \textbf{$\uparrow$} & \textbf{$\downarrow$} & \textbf{$\uparrow$} & \\
\midrule
\multicolumn{9}{c}{\textbf{MT-Bench}} \\
\midrule
Best-answer & 4.406 & 4.506 & 4.738 & 4.731 & 5.056 & 4.675 & \underline{5.088} & \textbf{5.125} \\
Random & \underline{4.470} & 4.469 & 4.688 & 4.695 & \textbf{4.785} & 4.500 & 4.581 & 4.613 \\
Top5-random & 4.435 & 4.681 & 4.870 & 4.788 & \underline{5.008} & 4.619 & 4.956 &  \textbf{5.048}\\
\midrule
\multicolumn{9}{c}{\textbf{Arena-Hard}} \\
\midrule
Best-answer & 75.3\scriptsize(-2.0, 1.6) & 78.6\scriptsize(-1.9, 2.1) & 76.8\scriptsize(-1.6, 1.7) & 81.8\scriptsize(-1.8, 1.2) & \textbf{83.3\scriptsize(-1.8, 1.7)} & 80.0\scriptsize(-1.5, 1.6) & \underline{82.3\scriptsize(-1.6, 2.2)} & 80.6\scriptsize(-2.4, 1,6)\\
Random & 74.5\scriptsize(-1.1, 1.2) & 78.5\scriptsize(-1.6, 1.3) & \textbf{80.4\scriptsize(-1.0, 1.5)} & 79.0\scriptsize(-1.3, 1.4) & 80.6\scriptsize(-1.6, 1.6) & 76.2\scriptsize(-0.8, 1.6) & 77.0\scriptsize(-1.0, 1.8) & 71.9\scriptsize(-1.7, 1.7)\\
Top5-random & 73.7\scriptsize(-1.2, 1.8) & 75.9\scriptsize(-1.6, 1.5) & 76.8\scriptsize(-1.2, 1.4) & \textbf{82.0\scriptsize(-1.3, 1.2)} & \underline{80.0\scriptsize(-0.7, 1.3)} & 75.0\scriptsize(-4.4, 5.8) & 76.9\scriptsize(-1.4, 1.6) & 76.6\scriptsize(-1.6, 1.5)\\
\bottomrule[1.5pt]
\end{tabular}
}
\label{fft - llama 3b instruct - response selection ablation - swp}
\end{table*}

\begin{table*}[!t]
\centering
\caption{Performance comparison of full fine-tuned Llama3.2-3b-base/instruct and Qwen2.5-3b-base/instruct models with different data selection strategies. The best and second results are in \textbf{bold} and \underline{underline}.}
\vspace{-1em}
\resizebox{\textwidth}{!}{ 
\begin{tabular}{cccccccccc}
\toprule[1.5pt]
\multirow{2}{*}{\textbf{Benchmark}} & \multirow{2}{*}{\textbf{Base}} & \multicolumn{3}{c}{\textbf{Baselines}} & \multicolumn{4}{c}{\textbf{Our Metrics}} \\
\cmidrule(lr){3-5} \cmidrule(lr){6-9}
 & &  \textbf{Random} & \textbf{Tags} & \textbf{IFD} & \textbf{Difficulty} & \textbf{Separability} & \textbf{Stability} & \textbf{Multi}\\
\midrule
\multicolumn{9}{c}{\textbf{Llama3.2-3b-base}} \\
\midrule
MT-Bench & 4.302 & 4.406 & 4.562 & 3.962 & 4.738 & 5.056 & \underline{5.088} & \textbf{5.125} \\
Arena-Hard & 50.0\scriptsize (-0.0, 0.0) & 75.3\scriptsize(-2.0, 1.6) & 77.3\scriptsize(-1.1, 1.2) & 77.6\scriptsize(-1.6, 1.6) & 76.8\scriptsize(-1.6, 1.7) & \textbf{83.3\scriptsize(-1.8, 1.7)} & 78.3\scriptsize(-1.6, 2.2) & \underline{80.6\scriptsize(-2.4, 1.6)}\\
\midrule
\multicolumn{9}{c}{\textbf{Llama3.2-3b-instruct}} \\
\midrule
MT-Bench & 6.200 & 6.356 & 6.393 & 6.243 & \underline{6.648} & 6.581 & 6.625 & \textbf{7.103} \\
Arena-Hard & 74.4\scriptsize (-1.0, 1.5) & 74.8\scriptsize(-1.5, 1.6) & \underline{81.6\scriptsize(-0.2, 0.2)} & 78.4\scriptsize(-1.7, 1.5) & 80.5\scriptsize(-0.9, 1.3) & 77.9\scriptsize(-1.5, 1.7) & 77.4\scriptsize(-1.5, 1.1) & \textbf{85.5\scriptsize(-0.8, 1.1)} \\
\midrule
\multicolumn{9}{c}{\textbf{Qwen2.5-3b-base}} \\
\midrule
MT-Bench & 6.043 & 6.500 & \underline{6.818} & 5.825 & 6.613 & \textbf{7.075} & 6.681 & 6.625\\
Arena-Hard & 69.0\scriptsize (-2.2, 1.6) & 72.9\scriptsize(-2.2, 1.9) & \underline{79.3\scriptsize(-2.2, 1.9)} & 74.5\scriptsize(-1.5, 1.5) & 73.8\scriptsize(-2.5, 1.8) & 74.1\scriptsize(-1.6, 2.4) & 76.8\scriptsize(-1.8, 1.8) &\textbf{ 79.9\scriptsize(-1.6,1.8)} \\
\midrule
\multicolumn{9}{c}{\textbf{Qwen2.5-3b-instruct}} \\
\midrule
MT-Bench & 7.138 & 6.793 & 6.818 & 6.731 & 7.182 & \underline{7.269} & \textbf{7.294} & 7.131 \\
Arena-Hard & 81.6\scriptsize (-1.8, 1.4) & 78.2\scriptsize(-1.7, 2.0) & 82.0\scriptsize(-2.4, 1.6) & 80.4\scriptsize(-1.3, 1.0) & 81.8\scriptsize(-1.6, 1.3) & \underline{83.7\scriptsize(-1.4, 1.2)} & 83.5\scriptsize(-1.4, 1.4) & \textbf{85.2\scriptsize(-1.2, 1.1)} \\
\bottomrule[1.5pt]
\end{tabular}
}
\label{tab:Full_Finetuning_Results}
\end{table*}

\subsection{Experiment Setups}
\paragraph{Datasets.} We conduct our experiments on Magpie-100K-Generator-Zoo\footnote{\url{https://huggingface.co/datasets/Magpie-Align/Magpie-100K-Generator-Zoo}} given that it directly matches our problem setting that contains answers from 19 models—Qwen2 \citep{Yang2024Qwen2TR}, Qwen2.5 \citep{yang2024qwen2}, Llama 3 \citep{dubey2024llama}, Llama 3.1 \citep{dubey2024llama}, Gemma 2 \citep{team2024gemma}, Phi-3 \citep{abdin2024phi} families and GPT-4 \citep{achiam2023gpt}—and their reward scores from three state-of-the-art reward models from RewardBench \citep{lambert2024rewardbench}: ArmoRM-Llama3-8B-v0.1 \citep{wang2024interpretable},  Skywork-Reward-Llama-3.1-8B \citep{liu2024skywork}, and Skywork-Reward-Gemma-2-27B \citep{liu2024skywork}.

\paragraph{Evaluation.} To evaluate the instruction-following capabilities, we use two widely-used instruction-following benchmarks: MT-Bench \citep{zheng2023judging} and Arena-Hard \citep{li2024crowdsourced}. Both benchmarks mainly leverage LLM-as-a-Judge \citep{zheng2023judging} for evaluation, while MT-Bench leverage 1-10 rating scoring and Arena-Hard leverage direct pairwise comparison and finally provide a leaderboard with one model as anchor-points. In our experiments, we set the base model (\emph{i.e.}, LLaMA-3.2-3B-base) as the anchor point for models for arena battles. We unify the LLM-as-a-Judge model in both benchmarks as DeepSeek-V3 \citep{liu2024deepseek} through official API\footnote{\url{https://platform.deepseek.com/}} and Together API\footnote{\url{https://api.together.ai/}} given its high performance on natural language generation tasks. Thanks to the unified judge model, we additionally report the \textbf{Average Performance (AP)} as a ranking computed by the ranking in MT-Bench and Arena-Hard. \textbf{Each experiment is conducted 3 times. The average results are reported to ensure the reliability and reproducibility.}

\paragraph{Base Models.}
Following \citep{xu2024stronger}, we consider four small models from different developers as student models, including base and instruct models—Qwen-2.5-3B, Qwen-2.5-3B-Instruct \citep{yang2024qwen2} and LLaMA-3.2-3B, LLaMA-3.2-3B-Instruct \citep{dubey2024llama}. We use 10 clusters for diversity preservation, and the multimetric setting uses $w = (1, 1, 2)$ for metric integration in the following experiments.

\paragraph{Baselines.} We include 7 baselines in our experiments. \textit{Random}, denotes a randomly selected instruction-answer set from the original dataset. We also compared two previous \emph{state-of-the-art} data selection method: Instag \citep{lu2023instag} , and IFD \citep{li2023quantity}. For rule-based method, we include \textit{Length} and \textit{Reward Score} \citep{Liu2023WhatMG}. More details are shown in Appendix \ref{Appendix: baseline intro}.

\paragraph{Instruction-Tuning Setups.}
We conduct our fine-tuning and evaluation on single A800 and A6000 servers. For fine-tuning, we use LLaMA-Factory \citep{zheng2024llamafactory}. For evaluation, we leverage the official codebase of MT-Bench\footnote{\url{https://github.com/lm-sys/FastChat/tree/main/fastchat/llm_judge}} and Arena-Hard\footnote{\url{https://github.com/lmarena/Arena-Hard-auto}} for automatic assessments. See Appendix \ref{Appendix: detailed experiment setups} for more details of experiment setups.

\begin{figure*}[!t]
    \centering
    \includegraphics[width=1\linewidth]{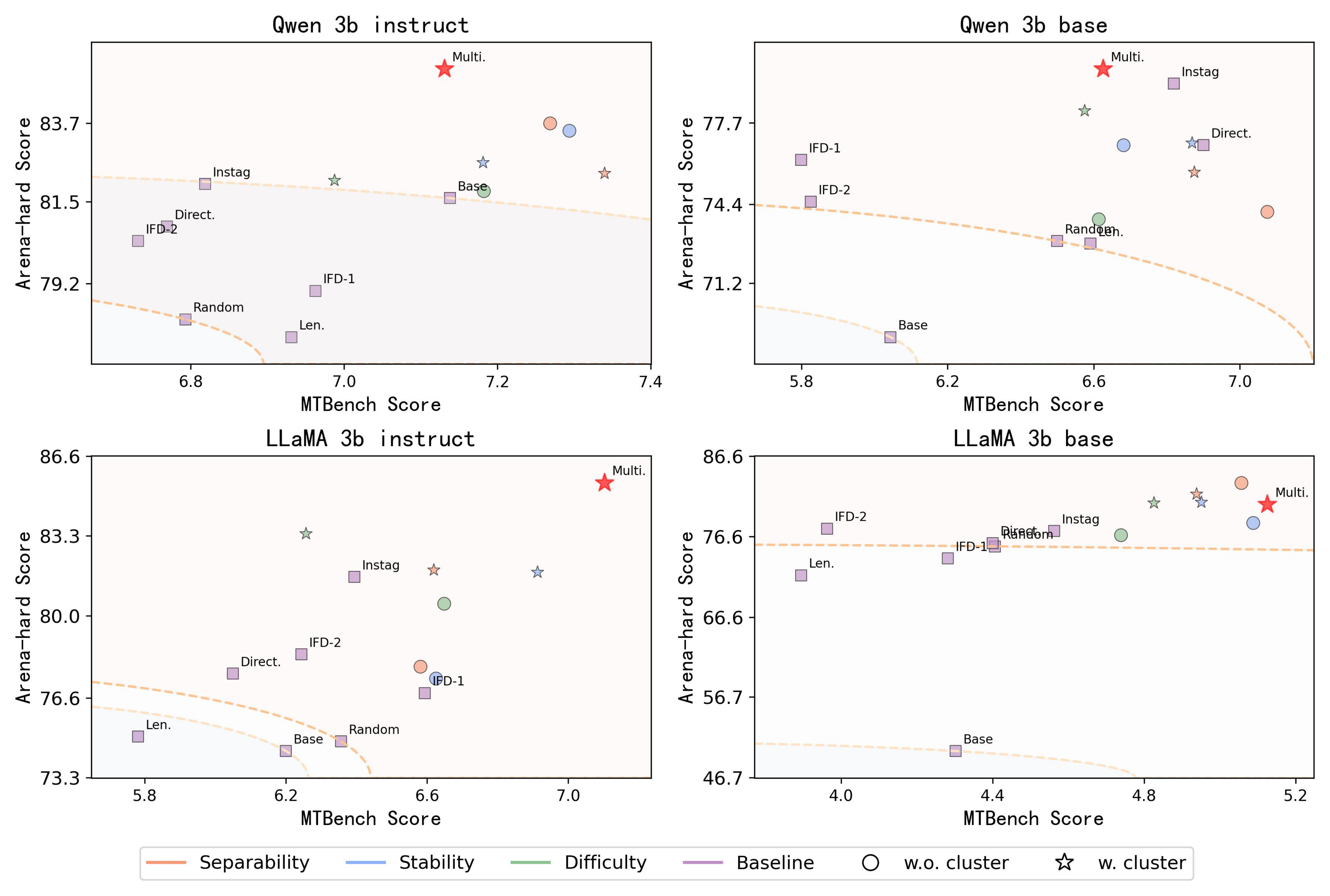}
    \vspace{-2em}
    \caption{Overall results demonstrate that our foundation metrics and \method consistently outperform baseline methods by a significant margin across FFT settings of four models, with particularly strong performance improvements on Llama-3b-instruct.}
    \vspace{-1em}
    \label{fig:overall results}
\end{figure*}

\subsection{Experiment Results.}

\paragraph{Three foundation metrics demonstrate effectiveness in selecting valuable samples.} As shown in Table \ref{fft - llama 3b instruct - response selection ablation - swp}, our three foundation metrics consistently identify valuable instruction samples across all response selection strategies. Models fine-tuned on \textit{Top-scored} samples consistently outperform \textit{Bottom-scored} samples, with \textit{Stability} exceed the most margin. We also explore the response selection strategies to build a foundation for following experiments. \textit{Best-answer} setting outperforms both \textit{Random} and \textit{Top5-random} approaches, indicating that responses with higher reward scores provide better quality data for distillation. This consistent performance across individual metrics establishes strong foundation for further improvements through integration. Therefore, we use \textit{top-scored} as the instruction selection and \textit{Best-answer} as the corresponding response for all experiments. 

\paragraph{\method achieves new \emph{state-of-the-art} performance on both benchmarks.} As shown in Table \ref{tab:Full_Finetuning_Results}, our approach significantly outperforms previous baselines across four models, demonstrating robust generalization. On Arena-Hard and MT-bench, \method with Llama-3.2-3b-instruct achieves scores of 85.5 and 7.103 respectively, surpassing the previous best results by 4.81\% and 11.1\%. For Qwen-2.5-3b-instruct, \method outperforms the strongest baseline by 3.90\%, validating our approach of post-training with high-quality instructions and model distillation. Even for base models, our foundation metrics and \method prove effective, notably improving Llama-3.2-3b's performance on MT-bench by 12.3\%.

\paragraph{\method performs robust on various fine-tuning methods.} Beyond demonstrating superior performance on standard benchmarks, the proposed metrics are further evaluated for robustness across a range of fine-tuning methodologies. Table~\ref{fft - llama 3b instruct - response selection ablation - swp} reveals consistent and stable performance of the proposed metrics. This robustness across varying training paradigms highlights the generalizability of the metrics and suggests their applicability in a wider range of practical scenarios.

\subsection{Ablation Studies} 
We conduct ablation studies for each module in \method to provide a comprehensive analysis of our approach. Further experiments on fine-tuning with LoRA, other training recipes, and ablation study for reward scores are also presented in Appendix \ref{Appendix: additional experiment results}. 

\begin{figure}[!t]
    \centering
    \includegraphics[width=1\linewidth]{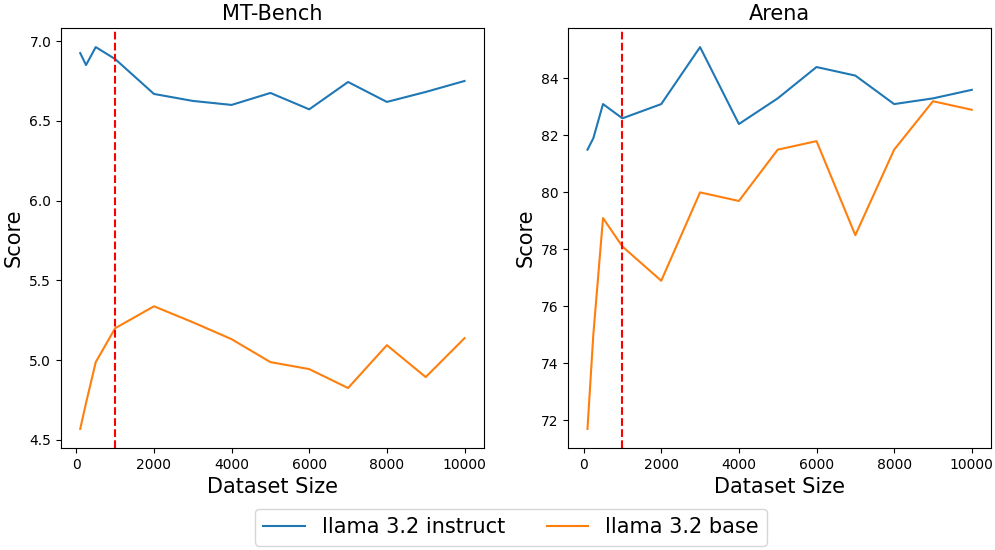}
    \vspace{-2em}
    \caption{Results show that small elite datasets behaves on par with a large dataset, corresponding to the experiment results in \citep{cao2023instruction}. Our implementation (line in \textcolor{red}{Red}) achieves reasonably good results.}
    \vspace{-1em}
    \label{fig:dataset_size}
\end{figure}

\paragraph{Dataset Size.} 
\citet{cao2023instruction} suggest that selecting concise subsets from all datasets can yield competitive results. Building on this insight, we collected 1$k$ instruction-response pairs for each setting in our main experiments. Additional experiments across various dataset sizes further support this finding, as the results in Figure~\ref{fig:dataset_size} show that small, high-quality datasets perform on par with larger datasets. This underscores the importance of data quality over sheer quantity in instruction tuning \cite{chen2023alpagasus,zhou2024lima}.

\paragraph{Metric Coefficient Combination.} Merging different metrics tend to achieve a better performance in synthetic data selection \citep{xu2024stronger,Liu2023WhatMG}. Our experiments follow this recipe to explore various coefficient combinations to determine the optimal balance for creating high-quality, robust datasets. Table \ref{main - fft compare - our metrics + cluster + multimetric - llama 3b instruct - swp} details the process of optimizing the weights assigned to different metrics when evaluating dataset quality.  Fine-tuning on subset selected by $w = (1, 1, 2)$ consistently yielded superior results compared to other tested combinations among 3/4 models in Tables \ref{fft compare - our metrics + cluster + multimetric - llama 3b instruct - swp}, \ref{fft compare - our metrics + cluster + multimetric - qwen 2.5 instruct}, and \ref{fft compare - our metrics + cluster + multimetric - llama3.2 base} in Appendix. 

\paragraph{Number of Clusters.} Clustering's impact on dataset quality was investigated by varying the number of clusters during dataset selection (see Table \ref{main - fft - llama 3b instruct - cluster ablation - swp}). While more cluster shows higher performance on \textit{Random} setting, no strong positive correlation on our metrics and \method between cluster count and quality. On the other hand, corresponding with previous research \citep{bukharin2023data, wang2024diversity}, data selection after clustering outperformed those constructed without clustering, highlighting the importance of enhancing robustness by the clustering process.

\paragraph{Response Generation Strategy.} 
The response selection strategy significantly impacts the fine-tuned LLM’s generation quality. As shown in Table \ref{fft - llama 3b instruct - response selection ablation - swp}, the best-answer strategy performs noticeably better than other approaches, underscoring the importance of high-quality responses within the dataset. We contend that the difficulty metric’s independence from response strategy stems from the fact that the core challenges of these instructions are intrinsically linked to the complexity of the tasks themselves, rather than the method used to formulate responses. For example, a particularly demanding instruction might require the model to synthesize knowledge from multiple domains, reason through abstract concepts, or produce detailed, contextually nuanced outputs \citep{shah2025aiassistedgenerationdifficultmath, rein2023gpqa}. Such requirements remain consistent, regardless of the response generation strategy employed.

\begin{table}[!t]
\centering
    \small
    \renewcommand{\arraystretch}{1.1}  
    \caption{Hyperparameter comparison of \method using Llama-3b-instruct models with varying cluster numbers.}
    \vspace{-1em}
    \resizebox{0.5\textwidth}{!}{
    \begin{tabular}{ccccc}
        \toprule
        \multicolumn{3}{c}{\textbf{Hyperparameter}} &\multirow{2}{*}{\textbf{MT-Bench}} & \multirow{2}{*}{\textbf{Arena-Hard}} \\
        \cmidrule(lr){1-3}
        \textbf{Diff.} & \textbf{Sep.} & \textbf{Stab.} & & \\
        \midrule
        1 & 1 & 1 & 6.913 & 81.8\scriptsize(-0.5, 0.8) \\
        1 & -1 & 1 & 6.625  & 84.2\scriptsize(-0.7, 1.0) \\
        1 & 1 & 2 & \textbf{7.103}  & \textbf{85.5}\scriptsize(-0.8, 1.1) \\
        1 & 1 & -1 &  6.650  & 82.7\scriptsize(-1.5, 1.4) \\
        1 & 1 & 1.5 & 6.850  & 84.7\scriptsize(-1.6, 1.3) \\
        1 & -1 & 1.5 & 6.781  & 83.0\scriptsize(-1.4, 1.4) \\
        -1 & -1 & 1 & 6.781  & 81.9\scriptsize(-1.5, 1.3) \\
        -1 & -1 & 2 & 6.838  & 84.8\scriptsize(-1.3, 1.2) \\
        -1 & -1 & 1.5 & 6.638  & 81.8\scriptsize(-1.3, 1.3) \\
        \bottomrule
    \end{tabular}
    }
    \label{main - fft compare - our metrics + cluster + multimetric - llama 3b instruct - swp}
\end{table}

\begin{table}[!t]
\centering
\setlength{\tabcolsep}{4pt} 
\caption{Performance comparison of FFT-version of Llama-3b-instruct on different coefficient combinations for multiple metrics with clustering.}
\vspace{-1em}
\resizebox{0.5\textwidth}{!}{
\begin{tabular}{ccccc}
\toprule[1.5pt]
\textbf{Benchmark} & \textbf{Random} & \textbf{Difficulty} & \textbf{Separability} & \textbf{Stability} \\
\midrule
\multicolumn{5}{c}{\textbf{10 clusters}} \\
\midrule
MT-Bench & 6.443 & 6.675 & 6.619 & \textbf{6.913} \\
Arena-Hard & 80.9 & \textbf{82.6} & 81.9 & 81.8 \\
Arena-Hard-95\%CI & (-1.3, 1.4) & (-1.2, 1.8) & (-1.7, 1.7) & (-1.5, 1.7)\\
\midrule
\multicolumn{5}{c}{\textbf{20 clusters}} \\
\midrule
MT-Bench & 6.607 & 6.615 & 6.591 & \textbf{6.686}  \\
Arena-Hard & 82.8 & 83.1 & \textbf{85.2}& 82.8  \\
Arena-Hard-95\%CI & (-1.2, 1.4) & (-1.1, 1.7) & (-1.3, 1.1) & (-1.4, 1.1)  \\
\midrule
\multicolumn{5}{c}{\textbf{30 clusters}} \\
\midrule
MT-Bench & 6.721& \textbf{6.737} & 6.725 & 6.562  \\
Arena-Hard & 83.2 & \textbf{84.9} & 83.3 & 83.8  \\
Arena-Hard-95\%CI & (-1.3, 1.1) & (-1.0, 1.1) & (-1.4, 1.4) & (-1.4, 1.2)  \\
\bottomrule[1.5pt]
\end{tabular}
}
\label{main - fft - llama 3b instruct - cluster ablation - swp}
\end{table}

\section{Conclusion}
\label{conclusion}

This paper presents novel metrics for synthetic instruction data selection based on Multi-LLM Wisdom, capturing the \textit{difficulty} of instructions from multiple perspectives through various LLMs' responses and their corresponding reward scores. We validate our hypothesis through the strong performance of individual metrics on both MT-Bench and Arena-Hard using FFT and LoRA fine-tuning on Llama-3.2-3b and Qwen-2.5-3b. By combining diversity enhancement through clustering with our proposed metrics, \method consistently outperforms \emph{state-of-the-art} data selection methods, establishing both new perspectives and a robust baseline for instruction tuning data selection.

\section*{Limitations}
\label{limitations}
\method exhibits notable progress in synthetic data selection tasks, yet some limitations remain. Our approach calculates selection metrics by employing responses from multiple model families and their associated reward scores, which may introduce reward model biases or reward hacking risks. While integrating these reward scores more seamlessly might improve robustness, doing so would require extra computational resources. Additionally, the experiments are performed on A800 and A6000 GPUs, and differences in hardware environments might introduce variability that could affect the reproducibility of our results.


\nocite{langley00}
\bibliography{custom}

\clearpage
\appendix

\section{Detailed Related Works}
\label{Appendix: detailed related works}

\paragraph{Instruction Tuning Data Selection.}
While LLMs like GPT-4 \citep{achiam2023gpt,openai_gpt4o_2024} and Llama-3 \citep{dubey2024llama} excel in natural language understanding and generation, their pre-training objectives often misalign with user goals for instruction-following tasks \citep{murthy2024evaluating,gao2024best,wen2024benchmarking}. Instruction tuning (or supervised fine-tuning) addresses this gap by refining LLMs on curated datasets of prompts and responses. Recent efforts like Vicuna \citep{Peng2023InstructionTW} and LIMA \citep{zhou2024lima} demonstrate high performance with a carefully selected small dataset, highlighting the growing importance of efficient instruction tuning and paving the way for aligning models with selected samples. This involves determining which instruction-response pairs to include in the training dataset and how to sample them effectively \citep{Albalak2024ASO}.

Three key metrics determine instruction data quality: \textit{Difficulty}, \textit{Quality}, and \textit{Diversity}. \textit{Difficulty}, focusing mainly on the question side, is considered more valuable for model learning \citep{liu2024selectit,lee2024instruction,wang2024survey}. IFD \citep{li2023quantity} pioneered the measurement of instruction-following difficulty for specific pairs, later enhanced by utilizing GPT-2 for efficient estimation in a weak-to-strong manner \citep{Li2024SuperfilteringWD}. \textit{Quality}, mainly addressing the response side, measures the helpfulness and safety of model responses, typically assessed using LLM evaluators \citep{chen2023alpagasus,chen2024mllm,Liu2024SelectITSI,ye2024justice}, reward models \citep{son2024llm,lambert2024rewardbench}, and gradient similarity search \citep{Xia2024LESSSI}. \textit{Diversity}, spanning both instruction and response aspects, plays a crucial role in covering various instruction formats and world knowledge, primarily improving model robustness \citep{bukharin2023data, wang2024diversity}.
Our work stands out by addressing all three key components in data selection, introducing novel approaches to measuring difficulty from multiple LLMs' responses and ultimately enhancing model performance.

\paragraph{Data Synthesis for Instruction Tuning.}
While the development of LLMs initially relied on human-curated instruction datasets for instruction tuning \citep{zheng2023lmsys,Zhao2024WildChat1C,Lightman2023LetsVS}, this approach proved time-consuming and labor-intensive, particularly as the complexity and scope of target tasks increased \citep{Demrozi2023ACR,Wang2021WantTR}. Consequently, researchers began exploring the use of frontier LLMs to generate synthetic instruction datasets, aiming to both address these scalability challenges  \citep{Ding2023EnhancingCL,Chen2023ShareGPT4VIL,Chen2024ShareGPT4VideoIV} and leverage  models' advanced capabilities in developing next-generation foundation models \citep{Burns2023WeaktoStrongGE, Li2024SuperfilteringWD,Charikar2024QuantifyingTG}.
Early approaches \citep{Xu2023WizardLMEL,Wang2024CodecLMAL,Zhou2024EvaluatingTV,Luo2023WizardCoderEC} focused on leveraging LLMs to generate synthetic instructions through a subset of human-annotated seed instructions \citep{chen2023alpagasus,Wang2023HowFC}, and further enhanced by few-shot \citep{Li2024SyntheticD} and attribute-guided prompting \citep{yu2023large,wu2024unigen,huang2024datagen}. A parallel line of research explored summarizing world knowledge to create more diverse synthetic datasets, aiming to maximize the coverage of different domains and task types \citep{Cui2023UltraFeedbackBL,Li2024SyntheticD}. Recent advancements have further streamlined this process by utilizing instructions directly from pre-trained LLMs with simple prompt templates \citep{Xu2024MagpieAD,Chen2024GenQAGM,Zhang2024ProVisionPS}, significantly reducing the required custom design from human effort.
While existing work has primarily focused on generating extensive, diverse, and high-quality datasets—often scaling to 100,000 examples or more—this approach introduces challenges in terms of computational efficiency and training resource requirements \citep{Li2024SCAREI,Dubois2024LengthControlledAA}. 

\paragraph{Deriving Crowded Wisdom from Multi-LLM.} Single LLM's response to a question face limitations in its representation of data (particularly cutting-edge knowledge) \citep{lazaridou2021mind,dhingra2022time,kasai2023realtime}, skills (as no single LLM is universally optimal \textit{empirically}) \citep{sun2022paradigm,liang2022holistic,chen2024interleaved}, and diverse perspectives \citep{Feng2025WhenOL}. Previous work has demonstrated that \textit{online} multi-LLM wisdom (also known as compositional agent frameworks \citep{gupta2023visual}) tends to outperform single models across various domains, providing more comprehensive and reflective solution on complex downstream tasks \citep{wang2024mixture,hong2023metagpt,wu2023autogen,li2023camel,ouyang2025nvagent,gui2025uicopilot}. \textit{Offline} crowded wisdom, where data are pre-collected rather than real-time inference, also show potential in model alignment \citep{gallego2024refined,rafailov2023direct,meng2025simpo} and benchmark construction \citep{ni2024mixeval,ni2024mixeval}. In this paper, we pioneer the use of \textit{offline} multi-LLM wisdom for instruction data selection by utilizing these LLMs' responses and their reward Score as \textit{reflections} to measure instruction-response pairs' \textit{Difficulty} and \textit{Quality}.

\section{Detailed Experiment Setups}
\label{Appendix: detailed experiment setups}

\subsection{Models \& Benchmarks \& Datasets Introduction}
\paragraph{Models.} In our study, the synthetic instruction dataset used for data selection consists of 19 response generators across 6 model families. These families include Qwen2 \citep{Yang2024Qwen2TR}, Qwen2.5 \cite{yang2024qwen2}, LLaMA 3 \cite{dubey2024llama}, LLaMA 3.1 \cite{dubey2024llama}, Gemma 2 \cite{team2024gemma}, and Phi-3 \cite{abdin2024phi}. In our experiments, we perform supervised fine-tuning on the LLaMA3.2-3B-base/instruct \cite{dubey2024llama} and Qwen-2.5-3b-base/instruct \cite{yang2024qwen2} models using the selected 1K datasets. A comprehensive overview of the models used in our study is presented in Table~\ref{tab:models_used_in_our_study}.

\begin{table*}[ht]
\centering
\caption{Overview of 22 models used in our study.}
\label{tab:models_used_in_our_study}
\begin{tabular}{l c c c}
\hline
\textbf{Model Family} & \textbf{Release Date} & \textbf{Model ID} & \textbf{Size} \\ \hline
                           &           & Qwen2-1.5B-Instruct  & 1.5B \\
       Qwen2               & Jun, 2024 & Qwen2-7B-Instruct    & 7B   \\
\citep{Yang2024Qwen2TR}   &           & Qwen2-72B-Instruct   & 72B  \\ \hline
                           &           & Qwen2.5-3B  & 3B   \\
                           &            & Qwen2.5-3B-Instruct  & 3B   \\
       Qwen2.5             &            & Qwen2.5-7B-Instruct  & 7B   \\
\citep{yang2024qwen2}     & Sept, 2024 & Qwen2.5-14B-Instruct & 14B  \\
                           &            & Qwen2.5-32B-Instruct & 32B  \\
                           &            & Qwen2.5-72B-Instruct & 72B  \\ \hline
    Llama 3                & Apr, 2024  & Llama-3-8B-Instruct  & 8B   \\
  \citep{dubey2024llama}  &            & Llama-3-70B-Instruct & 70B  \\ \hline
                           &           & Llama-3.1-8B-Instruct & 8B   \\
   Llama 3.1               & Jul, 2024  & Llama-3.1-70B-Instruct & 70B \\
\citep{dubey2024llama}    &            & Llama-3.1-405B-Instruct & 405B \\ \hline
    Llama 3.2              & Jul, 2024  & Llama-3.2-3B & 3B   \\
 \citep{dubey2024llama}   &            & Llama-3.2-3B-Instruct & 3B \\ \hline
                           &             & Gemma-2-2B-it        & 2B   \\
    Gemma 2                &  Jun, 2024  & Gemma-2-9B-it        & 9B   \\
   \citep{team2024gemma}  &            & Gemma-2-27B-it       & 27B  \\ \hline
                           &             & Phi-3-mini-128k-instruct & 3.5B \\
   Phi-3                   & Jun, 2024   & Phi-3-small-128k-instruct & 7B   \\
 \citep{abdin2024phi}     &            & Phi-3-medium-128k-instruct & 14B  \\ \hline
\end{tabular}
\end{table*}

\paragraph{Benchmarks.} In order to evaluate the instruction-following capabilities of the models, we use two widely-used 
instruction-following benchmarks: MT-Bench and Arena-Hard in our study.

\paragraph{MT-Bench \citep{zheng2023judging}.} MT-bench is a collection of open-ended questions designed to evaluate a chatbot’s performance in multi-turn conversations and its ability to follow instructions—two critical factors in aligning with human preferences. It consists of 80 high-quality multi-turn questions, which are divided into 8 categories: writing, roleplay, extraction, reasoning, mathematics, coding, knowledge I (STEM), and knowledge II (humanities/social sciences). Each category contains 10 questions. This framework provides a robust tool for assessing the practical effectiveness of LLMs and their alignment with human preferences, through meticulously designed questions and evaluations conducted by human annotators.

\paragraph{Arena-Hard \citep{li2024crowdsourced}.} Arena-Hard is a benchmark consisting 500 challenging prompts curated by BenchBuilder. It extracts high-quality prompts from crowdsourced datasets like Chatbot Arena \citep{zheng2023judging} and WildChat-1M \citep{Zhao2024WildChat1C} without human intervention.The prompts are Scored and filtered based on seven key qualities, including specificity, domain knowledge, complexity, problem-solving, creativity, technical accuracy, and real-world applicability. This ensures that the prompts are challenging and capable of distinguishing between models. Unlike static benchmarks, Arena-Hard can be continuously updated to reflect the latest advancements in LLMs, avoiding the risk of becoming obsolete or leaking test data.

\paragraph{Datasets.} In this paper, we conduct our experiments on Magpie-100K-Generator-Zoo\cite{xu2024stronger} because it provides a sufficiently large quantity of high-quality instruction fine-tuning data. It is a subset sampled from the MagpieAir-3M \cite{Xu2024MagpieAD} dataset, a large-scale instruction dataset. Magpie-100K contains 100,000 high-quality instructions, which are categorized into several types, including information seeking, mathematics, planning, coding and debugging, advice seeking, creative writing, reasoning, data analysis, brainstorming, editing, role-playing, and more.Each instruction has responses from 19 models across 6 model families—and their reward scores form 3 reward models. The diversity of these instructions ensures that the dataset covers a wide range of scenarios and tasks, making it suitable for instruction tuning of LLMs.

\subsection{Model Training Details}
\noindent
Table 2 demonstrates the detailed supervised fine-tuning (SFT) hyper-parameters. We perform experiments on a server with eight NVIDIA A800-SXM4-80GB GPUs, two Intel Xeon Platinum 8358P 64-Core Processor, and 1024 GB of RAM. These experiments were conducted using LLaMA-Factory~\cite{zheng2024llamafactory}.

\begin{table}[ht]
\centering
\caption{This table includes the hyper-parameters for supervised fine-tuning.}
\vspace{-1em}
\begin{tabular}{l c}
\toprule[1.5pt]
\textbf{Hyper-parameter} & \textbf{Value} \\
\midrule
Learning Rate & $1 \times 10^{-5}$ \\
Number of Epochs & 3 \\
Per-device Batch Size & 1 \\
Gradient Accumulation Steps & 2 \\
Optimizer & Adamw \\
Learning Rate Scheduler & cosine \\
Warmup Steps & 150 \\
Max Sequence Length & 2048 \\
\bottomrule[1.5pt]
\end{tabular}
\end{table}

\subsection{Baseline Introduction}
\label{Appendix: baseline intro}
We present five baseline methods for comparison in our study. For each baseline, we describe its implementation details and rationale for inclusion.

\paragraph{Length-Based Filtering \citep{kwon2024instructcmp}.}
The Length method filters instructions based on their token count. We use the LLaMA 3.2 3B Instruction tokenizer to compute the number of tokens in each instruction. Instructions that meet the predefined length criteria are selected for further processing.

\paragraph{Instag-Based Selection \cite{lu2023instag}.}
The Instag method incorporates instruction tagging to examine the supervised fine-tuning process of LLMs. Our implementation involves the following steps:  
First, we leverage DeepSeek's API to obtain the true labels for the instructions.  
Next, instructions are grouped according to their respective labels.  
Then, we compute the complexity and diversity within each group.  
Finally, we select a subset of instructions that demonstrate the most desirable characteristics.

\paragraph{Direct Score Filtering.}
The Direct Score method is inspired by the work of \cite{chen2023alpagasus}, which proposes a scoring mechanism for instruction selection. 
We use the same prompt templates as the original paper.
Instead of the original scoring model, we use DeepSeek for scoring, ensuring consistency with our other experimental setups.
We select the top 1,000 instructions based on their scores.

\paragraph{Instruction Filtering by IFD.}
This approach builds on the work of \cite{li2023quantity}, which introduces self-guided data selection as a means of improving instruction tuning. We use the open-source implementation from Cherry LLM and employ a three-step process: 1) train a Pre-Experienced Model to establish prior knowledge, 2) calculate IFD (Instruction Filtering Degree) with the Pre-Experienced Model, and 3) filter the dataset based on IFD scores to retain high-quality instructions.

To assess the effectiveness of IFD, we consider two variants:  
1) IFD (with pre): This version utilizes a trained Pre-Experienced Model to compute IFD.  
2) IFD (no pre): This version computes IFD directly using the model being trained.

\paragraph{Random Sampling.}
The Random baseline selects a random subset of 1,000 instructions. Additionally, for each instruction, we randomly select one of its 19 possible responses, ensuring that instruction-response pairs are fully randomized.

\section{Additional Experiment Results}
\label{Appendix: additional experiment results}

\subsection{Dataset Size Ablation Details} Tables \ref{llama 3b instruct different data size - difficulty} and \ref{llama 3b base different data size - difficulty} detail the training loss, evaluation loss, and scores of Llama3.2-3b-base/instruct fine-tuned on different dataset sizes when selected with the difficulty metric. The data clearly shows a rapid increase in accuracy in when increasing the dataset sizes up to 0.5$k$ to 1$k$, and marginal increases afterwards. This highlights the importance of data quality over sheer quantity in instruction tuning.

\subsection{\method Performance on LoRA} Tables \ref{lora compare - our metrics - data selection strategy - swp} and \ref{lora compare - baselines - data selection strategy} detail the performance of \method and various baselines combined with LoRA fine-tuning. \method generally outperforms the baseline dataset selection methods on LoRA. However, more instability is found in LoRA training due to its limited learning capability compared with full fine-tuning.

\subsection{\method Performance on Full Fine-tuning} Tables \ref{fft compare - our metrics - data selection strategy - swp} and \ref{fft compare - baselines - data selection strategy} detail the performance of \method and various baselines combined with Full fine-tuning.

\subsection{Foundation Metric with Clustering Performance} Table \ref{fft compare - our metrics + cluster - data selection strategy} details the performance of our foundation metric combined with clustering strategy.

\subsection{\method Integrated Metric Performance on Different Coefficient Combinations} Tables \ref{fft compare - our metrics + cluster + multimetric - llama 3b instruct - swp}, \ref{fft compare - our metrics + cluster + multimetric - qwen 2.5 instruct}, and \ref{fft compare - our metrics + cluster + multimetric - llama3.2 base} detail the performance of our Integrated metric performance on 9 sets of coefficients. $w = (1, 1, 2)$ stands out as stable coefficients among all other combinations.

\subsection{\method Performance on Different Fine-tuning Methods} Table \ref{training method comparison - llama 3b instruct - swp} details the performance of \method on SFT \citep{ouyang2022training}, DPO \cite{rafailov2023direct}, SimPO \citep{meng2025simpo}, and ORPO \citep{hong-etal-2024-orpo}. Data reveals consistent and stable performance our proposed metrics, while SimPO performs best on all scenarios.

\subsection{\method Performance on Different Reward Models} 
Table \ref{lora - llama 3b instruct - reward model ablation - swp} presents the performance of \method on various reward models, emphasizing the significant impact that reward models have on fine-tuned model performance. The results reveal a nuanced landscape in which the strengths of different reward models are distributed across various performance metrics. This scattered performance underscores the importance of careful reward model selection and highlights the high variance among current LLM-based reward models. Consequently, further research into more robust reward models for LLMs is crucial.

\section{Case Study}
We present the top-5 instruction-response pairs generated by our fine-tuned models, as selected based on our foundation metrics in Tables~\ref{tab:instruction_response_difficulty}, \ref{tab:instruction_response_seperability}, \ref{tab:instruction_response_stability}, and \method in Table~\ref{tab:instruction_response_crowd}. We also present an example for fine-tuned in Figure~\ref{fig: example}.

\begin{table*}[h]
    \centering
    \small
    \setlength{\tabcolsep}{10pt}  
    \renewcommand{\arraystretch}{1.2}
    \caption{Performance comparison of Llama-3b-instruct with different sizes of difficulty-based selected data.}
    \vspace{-1em}
    \begin{tabular}{l@{\hspace{15pt}}ccccccc}
        \toprule
        \multirow{2}{*}{\textbf{Data Size}}& \multirow{2}{*}{\textbf{Train Loss}} & \multirow{2}{*}{\textbf{Eval. Loss}} & \multicolumn{2}{c}{\textbf{MT-Bench}} & \multicolumn{3}{c}{\textbf{Arena-Hard}} \\
        \cmidrule(r){4-5} \cmidrule(l){6-8}
        & & & \textbf{Score} & \textbf{Avg. Tokens} & \textbf{Score} & \textbf{95\% CI} & \textbf{Avg. Tokens} \\
        \midrule
        0.25$k$ & 0.418 & 0.951 & 6.850 & \textbf{301} & 81.9 & $(-1.2, 1.5)$ & 275 \\
        0.5$k$ & 0.406 & 1.004 & \textbf{6.962} & 276 & 83.1 & $(-1.0, 1.1)$ & 275 \\
        1$k$ & 0.407 & 0.942 & 6.887 & 271 & 82.6 & $(-1.5, 1.2)$ & 273 \\
        2$k$ & 0.405 & 0.929 & 6.668 & 301 & 83.1 & $(-1.0, 1.4)$ & 273 \\
        3$k$ & 0.415 & 0.871 & 6.625 & 304 & 85.1 & $(-1.3, 1.3)$ & \textbf{276} \\
        4$k$ & 0.413 & 0.869 & 6.600 & 279 & 82.4 & $(-1.1, 1.7)$ & 268 \\
        5$k$ & 0.415 & 0.867 & 6.675 & 295 & 83.3 & $(-0.7, 1.4)$ & 272 \\
        6$k$ & 0.414 & 0.857 & 6.572 & 282 & \textbf{84.4} & $(-1.1, 1.3)$ & 265 \\
        7$k$ & 0.413 & 0.848 & 6.743 & 286 & 84.1 & $(-0.9, 1.2)$ & 266 \\
        8$k$ & 0.411 & 0.836 & 6.618 & 275 & 83.1 & $(-1.1, 1.6)$ & 268 \\
        9$k$ & 0.411 & 0.822 & 6.681 & 274 & 83.3 & $(-1.3, 1.5)$ & 269 \\
        10$k$ & 0.409 & 0.828 & 6.750 & 279 & 83.6 & $(-0.8, 1.7)$ & 266 \\
        \bottomrule
    \end{tabular}
    \label{llama 3b instruct different data size - difficulty}
\end{table*}

\begin{table*}[h]
    \centering
    \small
    \setlength{\tabcolsep}{10pt}  
    \renewcommand{\arraystretch}{1.2}
    \caption{Performance comparison of Llama-3b with different sizes of difficulty-based selected data.}
    \vspace{-1em}
    \begin{tabular}{l@{\hspace{15pt}}ccccccc}
        \toprule
        \multirow{2}{*}{\textbf{Data Size}} & \multirow{2}{*}{\textbf{Train Loss}} & \multirow{2}{*}{\textbf{Eval. Loss}} & \multicolumn{2}{c}{\textbf{MT-Bench}} & \multicolumn{3}{c}{\textbf{Arena-Hard}} \\
        \cmidrule(r){4-5} \cmidrule(l){6-8}
        & & & \textbf{Score} & \textbf{Avg. Tokens} & \textbf{Score} & \textbf{95\% CI} & \textbf{Avg. Tokens} \\
        \midrule
        0.25$k$ & 0.567 & 1.138 & 4.731 & \textbf{492} & 75.0 & $(-1.1, 2.1)$ & 289 \\
        0.5$k$ & 0.544 & 1.161 & 4.987 & 392 & 79.1 & $(-1.0, 1.7)$ & 289 \\
        1$k$ & 0.539 & 1.123 & 5.200 & 325 & 78.1 & $(-1.4, 1.5)$ & 289 \\
        2$k$ & 0.534 & 1.094 & \textbf{5.337} & 309 & 76.9 & $(-1.4, 2.2)$ & \textbf{290} \\
        3$k$ & 0.537 & 1.046 & 5.237 & 286 & 80.0 & $(-1.6, 1.6)$ & 289 \\
        4$k$ & 0.535 & 1.031 & 5.131 & 287 & 79.7 & $(-1.3, 1.5)$ & 289 \\
        5$k$ & 0.534 & 1.022 & 4.987 & 271 & 81.5 & $(-1.0, 1.5)$ & 289 \\
        6$k$ & 0.531 & 1.019 & 4.943 & 251 & 81.8 & $(-1.3, 1.5)$ & \textbf{290} \\
        7$k$ & 0.529 & 1.004 & 4.825 & 218 & 78.5 & $(-1.2, 1.7)$ & 289 \\
        8$k$ & 0.526 & 0.990 & 5.093 & 278 & 81.5 & $(-1.1, 1.3)$ & 289 \\
        9$k$ & 0.519 & 0.982 & 4.893 & 245 & \textbf{83.2} & $(-1.5, 1.2)$ & 289 \\
        10$k$ & 0.517 & 0.983 & 5.137 & 270 & 82.9 & $(-1.0, 1.1)$ & 289 \\
        \bottomrule
    \end{tabular}
    \label{llama 3b base different data size - difficulty}
\end{table*}

\begin{table*}[htbp]
\centering
\setlength{\tabcolsep}{6pt}
\caption{Performance comparison of lora-version of Llama-3b-base/instruct and Qwen-3b-base/instruct models with different data selection strategies.}
\vspace{-1em}
\resizebox{\textwidth}{!}{
\begin{tabular}{cccccccc}
\toprule[1.5pt]
\multirow{2}{*}{\textbf{Benchmark}} & \multirow{2}{*}{\textbf{Base}} & \multicolumn{2}{c}{\textbf{Difficulty}} & \multicolumn{2}{c}{\textbf{Separability}} & \multicolumn{2}{c}{\textbf{Stability}} \\
& & \textbf{$\downarrow$} & \textbf{$\uparrow$} & \textbf{$\downarrow$} & \textbf{$\uparrow$} & \textbf{$\downarrow$} & \textbf{$\uparrow$} \\
\midrule
\multicolumn{8}{c}{\textbf{Llama3.2-3b-instruct}} \\
\midrule
MT-Bench & 6.200 & 6.456 & 6.688 & 6.100 & 6.725 & 6.131 & \textbf{6.866} \\
Arena-Hard & 74.4 & 69.6 & \textbf{76.8} & 69.4 & 72.9 & 69.8 & 74.6 \\
Arena-Hard-95\%CI &(-1.0, 1.5) & (-1.8,1.4) & (-1.5,1.9) & (-2.5,1.2) & (-1.6,1.5) & (-1.7,1.7) & (-1.7,2.0)\\ 
\midrule
\multicolumn{8}{c}{\textbf{Llama3.2-3b-base}} \\
\midrule
MT-Bench & 4.302 & 4.626 & \textbf{4.651} & 4.631 & 5.040 & 3.538 & 4.369 \\
Arena-Hard & 50.0 & 73.1 & 68.0 & \textbf{73.8} & 73.2 & 60.8 & 73.2 \\
Arena-Hard-95\%CI &(0.0,0.0) & (-1.8,1.6) & (-1.2,1.9) & (-1.2,1.8) & (-2.0,1.1) & (-1.7,1.2) & (-1.2,1.2)\\
\midrule
\multicolumn{8}{c}{\textbf{Qwen2.5-3b-instruct}} \\
\midrule
MT-Bench &\textbf{ 7.138} & 6.906 & 7.068 & 7.025 & 6.937 & 7.018 & 7.037 \\
Arena-Hard & \textbf{81.6} & 77.2 & 79.1 & 80.3 & 78.8 & 76.2 & 78.0 \\
Arena-Hard-95\%CI & (-1.8, 1.4) & (-1.9, 1.5) & (-2.1, 1.8) & (-1.9, 1.4) & (-1.2, 1.2) & (-1.7, 1.6) & (-1.8, 1.7) \\
\midrule
\multicolumn{8}{c}{\textbf{Qwen2.5-3b}} \\
\midrule
MT-Bench & 6.043 & 5.137 & \textbf{6.612} & 6.368 & 6.343 & 5.800 & 6.525 \\
Arena-Hard & 69.0 & \textbf{76.9} & 70.7 & 74.1 & 74.2 & 73.7 & 74.2 \\
Arena-Hard-95\%CI & (-2.2, 1.6) & (-2.0, 1.8) & (-1.8, 2.4) & (-1.8, 1.5) & (-2.1, 1.5) & (-2.0, 1.3) & (-1.8, 1.9) \\
\bottomrule[1.5pt]
\end{tabular}
}
\label{lora compare - our metrics - data selection strategy - swp}
\end{table*}

\begin{table*}[htbp]
\centering
\setlength{\tabcolsep}{4pt}
\caption{Performance comparison of lora-version of Llama-3b-base/instruct and Qwen-3b-base/instruct models with pre data selection strategies as baselines.}
\vspace{-1em}
\resizebox{\textwidth}{!}{
\begin{tabular}{ccccccccc}
\toprule[1.5pt]
\multirow{2}{*}{\textbf{Benchmark}} 
& \multirow{2}{*}{\textbf{Random}} 
& \multirow{2}{*}{\textbf{Tags}} 
& \multicolumn{2}{c}{\textbf{Direct-Score}} 
& \multicolumn{2}{c}{\textbf{Length}} 
& \multicolumn{2}{c}{\textbf{IFD}} \\
& & & \textbf{$\downarrow$} & \textbf{$\uparrow$} 
& \textbf{$\downarrow$} & \textbf{$\uparrow$} 
& \textbf{no\_pre} & \textbf{pre} \\
\midrule
\multicolumn{9}{c}{\textbf{Llama3.2-3b-instruct}} \\
\midrule
MT-Bench & 6.325 & 6.610 & 6.631 & 6.406 & 6.087 & 5.375 & 6.706 & \textbf{6.768} \\
Arena-Hard & 74.2 & 80.1 & 80.0 & 74.8 & 78.1 & 67.5 & \textbf{81.2} & 79.5 \\
Arena-Hard-95\%CI  &(-1.7, 1.3) & (-0.7, 0.7) &  (-1.4, 1.7) & (-1.1, 1.8) & (-3.4, 2.1) & (-1.4, 0.9) &  (-0.8, 1.5) & (-1.6, 1.8)\\
\midrule
\multicolumn{9}{c}{\textbf{Llama3.2-3b-base}} \\
\midrule
MT-Bench & 4.637 & 4.575 & \textbf{4.962} & 4.675 & 4.062 & 4.243 & 4.512 & 4.418 \\
Arena-Hard & 76.0 & 76.8 & 76.9 & 75.6 & 67.1 & 70.3 & 73.7 & \textbf{77.5} \\
Arena-Hard-95\%CI  &(-2.0, 1.6) & (-1.6, 1.8) & (-1.8, 1.7) & (-1.6, 1.4) & (-2.0, 2.0) & (-2.3, 2.2) &  (-1.5, 1.5) & (-1.8, 1.4)\\
\midrule
\multicolumn{9}{c}{\textbf{Qwen2.5-3b-instruct}} \\
\midrule
MT-Bench & 6.950 & 7.125 & 7.131 & \textbf{7.175} & 7.037 & 7.006 & 6.918 & 6.868 \\
Arena-Hard & 78.2 & 83.0 & 77.7 & 81.7 & 75.8 & 76.4 & 78.8 & \textbf{83.1} \\
Arena-Hard-95\%CI  &(-1.5, 1.8) & (-1.7, 2.1) &  (-1.6, 2.0) & (-1.7, 1.9) & (-2.0, 2.0) & (-1.4, 1.7) &  (-1.3, 1.2) & (-0.8, 1.0)\\
\midrule
\multicolumn{9}{c}{\textbf{Qwen2.5-3b-base}} \\
\midrule
MT-Bench & 5.887 & 5.616 & 5.417 & 5.750 & 3.981 & 5.637 & \textbf{6.427} & 5.861 \\
Arena-Hard & 76.6 & \textbf{83.8} & 79.3 & 76.5 & 74.3 & 70.4 & 79.7 & 82.2 \\
Arena-Hard-95\%CI  &(-1.7, 1.5) & (-1.3, 1.2) & (-1.8, 1.2) & (-2.0, 1.7) & (-1.8, 1.6) & (-1.6, 1.9) &  (-1.3, 1.0) & (-1.3, 1.0)\\
\bottomrule[1.5pt]
\end{tabular}
}
\label{lora compare - baselines - data selection strategy}
\end{table*}

\begin{table*}[htbp]
\centering
\setlength{\tabcolsep}{6pt}
\caption{Performance comparison of fft-version of Llama-3b-base/instruct and Qwen-3b-base/instruct models with different data selection strategies.}
\vspace{-1em}
\resizebox{\textwidth}{!}{
\begin{tabular}{cccccccc}
\toprule[1.5pt]
\multirow{2}{*}{\textbf{Benchmark}} & \multirow{2}{*}{\textbf{Base}} & \multicolumn{2}{c}{\textbf{Difficulty}} & \multicolumn{2}{c}{\textbf{Separability}} & \multicolumn{2}{c}{\textbf{Stablity}} \\
& & \textbf{$\downarrow$} & \textbf{$\uparrow$} & \textbf{$\downarrow$} & \textbf{$\uparrow$} & \textbf{$\downarrow$} & \textbf{$\uparrow$} \\
\midrule
\multicolumn{8}{c}{\textbf{Llama3.2-3b-instruct}} \\
\midrule
MT-Bench & 6.200 & 6.388 & \textbf{6.648} & 5.937 & 6.581 & 6.225 & 6.625 \\
Arena-Hard & 74.4 & 76.5 & \textbf{80.5} & 80.0 & 77.9 & 75.8 & 77.4 \\
Arena-Hard-95\%CI & (-1.0, 1.5) & (-1.6, 1.5) & (-0.9, 1.3) & (-1.3, 1.2) & (-1.5, 1.7) & (-1.3, 0.9) & (-1.5, 1.1) \\
\midrule
\multicolumn{8}{c}{\textbf{Llama3.2-3b-base}} \\
\midrule
MT-Bench & 4.302 & 4.506 & 4.738 & 4.731 & 5.056 & 4.675 & \textbf{5.088} \\
Arena-Hard & 50.0 & 78.6 & 76.8 & 81.8 & \textbf{83.3} & 80.0 & 78.3 \\
Arena-Hard-95\%CI & (0.0, 0.0) & (-1.9, 2.1) & (-1.6, 1.7) & (-1.8, 1.2) & (-1.8, 1.7) & (-1.5, 1.6) & (-1.6, 2.2) \\
\midrule
\multicolumn{8}{c}{\textbf{Qwen2.5-3b-instruct}} \\
\midrule
MT-Bench & 7.138 & 6.906 & 7.182 & 6.919 & 7.269 & 7.056 & \textbf{7.294} \\
Arena-Hard & 81.6 & 82.5 & 81.8 & 81.4 & \textbf{83.7} & 78.1 & 83.5 \\
Arena-Hard-95\%CI & (-1.8, 1.4) & (-1.8, 1.5) & (-1.6, 1.3) & (-1.7, 1.6) & (-1.4, 1.2) & (-1.2, 2.0) & (-1.4, 1.4) \\
\midrule
\multicolumn{8}{c}{\textbf{Qwen2.5-3b-base}} \\
\midrule
MT-Bench & 6.043 & 6.619 & 6.613 & 6.575 & \textbf{7.075} & 6.763 & 6.681 \\
Arena-Hard & 69.0 & \textbf{80.2} & 73.8 & 76.5 & 74.1 & 74.4 & 76.8 \\
Arena-Hard-95\%CI & (-2.2, 1.6) & (-1.7, 1.6) & (-2.5, 1.8) & (-1.8, 1.8) & (-1.6, 2.4) & (-1.5, 1.8) & (-1.8, 1.8) \\
\bottomrule[1.5pt]
\end{tabular}
}
\label{fft compare - our metrics - data selection strategy - swp}
\end{table*}

\begin{table*}[htbp]
\centering
\setlength{\tabcolsep}{4pt} 
\caption{Performance comparison of fft-version of Llama-3b-base/instruct and Qwen-3b-base/instruct models with pre data selection strategies as baselines.}
\vspace{-1em}
\resizebox{\textwidth}{!}{
\begin{tabular}{ccccccccc}
\toprule[1.5pt]
\multirow{2}{*}{\textbf{Benchmark}} 
& \multirow{2}{*}{\textbf{Random}} 
& \multirow{2}{*}{\textbf{Tags}} 
& \multicolumn{2}{c}{\textbf{Direct-Score}} 
& \multicolumn{2}{c}{\textbf{Length}} 
& \multicolumn{2}{c}{\textbf{IFD}} \\
& & & \textbf{$\downarrow$} & \textbf{$\uparrow$} 
& \textbf{$\downarrow$} & \textbf{$\uparrow$} 
& \textbf{no\_pre} & \textbf{pre} \\
\midrule
\multicolumn{9}{c}{\textbf{Llama3.2-3b-instruct}} \\
\midrule
MT-Bench & 6.356 & 6.393 & 6.068 & 6.050 & 5.612 & 5.781 & \textbf{6.593} & 6.243 \\
Arena-Hard & 74.8 & \textbf{81.6} & 76.9 & 77.6 & 72.9 & 75.0 & 76.8 & 78.4 \\
Arena-Hard-95\%CI  &(-1.5, 1.6) & (-0.2, -0.2) &  (-1.5, 2.0) & (-1.7, 1.9) & (-1.9, 1.9) & (-2.4, 2.0) &  (-1.2, 1.6) & (-1.7, 1.5)\\
\midrule
\multicolumn{9}{c}{\textbf{Llama3.2-3b-base}} \\
\midrule
MT-Bench & 4.406 & \textbf{4.562} & 4.131 & 4.400 & 3.393 & 3.893 & 4.281 & 3.962 \\
Arena-Hard & 75.3 & 77.3 & 72.7 & 75.8 & 59.4 & 71.8 & 73.9 & \textbf{77.6} \\
Arena-Hard-95\%CI  &(-2.0, 1.6) & (-1.1, 1.2) & (-2.4, 1.9) & (-1.4, 1.2) & (-1.1, 1.3) & (-1.0, 1.2) &  (-1.0, 1.6) & (-1.6, 1.6)\\
\midrule
\multicolumn{9}{c}{\textbf{Qwen2.5-3b-instruct}} \\
\midrule
MT-Bench & 6.793 & 6.818 & 6.506 & 6.768 & 5.881 & 6.931 & \textbf{6.962} & 6.731 \\
Arena-Hard & 78.2 & \textbf{82.0} & 81.2 & 80.8 & 75.6 & 77.7 & 79.0 & 80.4 \\
Arena-Hard-95\%CI  &(-1.7, 2.0) & (-2.4, 1.6) &  (-1.5, 1.8) & (-2.1, 1.7) & (-1.0, 1.2) & (-1.7, 1.7) &  (-1.0, 1.5) & (-1.3, 1.0)\\
\midrule
\multicolumn{9}{c}{\textbf{Qwen2.5-3b-base}} \\
\midrule
MT-Bench & 6.500 & 6.818 & 6.325 & \textbf{6.900} & 4.925 & 6.591 & 5.798 & 5.825 \\
Arena-Hard & 72.9 & \textbf{79.3} & 75.6 & 76.8 & 71.2 & 72.8 & 76.2 & 74.5 \\
Arena-Hard-95\%CI  &(-2.2, 1.9) & (-2.2, 1.9) & (-1.6, 2.1) & (-1.9, 1.9) & (-1.7, 1.4) & (-2.3, 1.9) &  (-1.4, 1.3) & (-1.5, 1.5)\\
\bottomrule[1.5pt]
\end{tabular}
}
\label{fft compare - baselines - data selection strategy}
\end{table*}

\begin{table*}[htbp]
\centering
\setlength{\tabcolsep}{4pt}
\caption{Performance comparison of cluster-chosen-data-fft-version of Llama-3b-base/instruct and Qwen-3b-base/instruct models with different data selection strategies.}
\vspace{-1em}
\resizebox{\textwidth}{!}{
\begin{tabular}{ccccccccc}
\toprule[1.5pt]
\multirow{2}{*}{\textbf{Benchmark}} 
& \multirow{2}{*}{\textbf{Base}} 
& \multirow{2}{*}{\textbf{Random}} 
& \multicolumn{2}{c}{\textbf{Difficulty}} 
& \multicolumn{2}{c}{\textbf{Separability}} 
& \multicolumn{2}{c}{\textbf{Stability}} \\
& & & \textbf{$\downarrow$} & \textbf{$\uparrow$} 
& \textbf{$\downarrow$} & \textbf{$\uparrow$} 
& \textbf{$\downarrow$} & \textbf{$\uparrow$} \\
\midrule
\multicolumn{9}{c}{\textbf{Llama3.2-3b-instruct}} \\
\midrule
MT-Bench & 6.200 & 6.743 & 6.256 & 6.675 & 6.094 & 6.619 & 6.275 & \textbf{6.913} \\
Arena-Hard & 74.4 & 80.9 & 81.4 & 82.6 & \textbf{84.8} & 81.9 & 80.0 & 81.8 \\
Arena-Hard-95\%CI  &(-1.0, 1.5) & (-1.3, 1.4) &  (-1.5, 2.0) & (-1.2, 1.8) & (-1.7, 1.4) & (-1.7, 1.7) &  (-2.0, 2.2) & (-1.5, 1.7)\\
\midrule
\multicolumn{9}{c}{\textbf{Llama3.2-3b-base}} \\
\midrule
MT-Bench & 4.302 & 4.869 & 4.825 & \textbf{5.000} & 4.813 & 4.938 & 4.800 & 4.950 \\
Arena-Hard & 50.0 & 79.2 & 80.8 & 79.5 & 80.8 & \textbf{81.9} & 80.6 & 80.9 \\
Arena-Hard-95\%CI & (0.0, 0.0) & (-0.9, 0.9) & (-1.2, 1.7) & (-1.7, 2.2) & (-2.0, 1.6) & (-1.5, 2.1) & (-1.9, 1.8) & (-2.0, 1.6) \\
\midrule
\multicolumn{9}{c}{\textbf{Qwen2.5-3b-instruct}} \\
\midrule
MT-Bench & 7.138 & 7.006 & 6.988 & 7.150 & 7.238 & \textbf{7.340} & 7.019 & 7.181 \\
Arena-Hard & 81.6 & 82.3 & 82.1 & \textbf{82.6} & 82.5 & 82.3 & 80.3 & \textbf{82.6} \\
Arena-Hard-95\%CI & (-1.8, 1.4) & (-1.0, 0.9) & (-1.6, 1.3) & (-1.9, 1.7) & (-2.1, 1.3) & (-1.0, 1.4) & (-1.5, 1.4) & (-1.4, 2.0) \\
\midrule
\multicolumn{9}{c}{\textbf{Qwen2.5-3b-base}} \\
\midrule
MT-Bench & 6.043 & \textbf{7.162} & 6.575 & 6.800 & 6.856 & 6.875 & 6.819 & 6.869 \\
Arena-Hard & 69.0 & 74.6 & 78.2 & \textbf{78.5} & 78.0 & 75.7 & 73.6 & 76.9 \\
Arena-Hard-95\%CI & (-2.2, 1.6) & (-0.7, 1.0) & (-1.9, 2.4) & (-1.6, 1.7) & (-1.7, 1.8) & (-2.2, 2.1) & (-1.8, 1.8) & (-2.1, 1.6) \\
\bottomrule[1.5pt]
\end{tabular}
}
\label{fft compare - our metrics + cluster - data selection strategy}
\end{table*}

\begin{table*}[h]
\centering
    \small
    \setlength{\tabcolsep}{10pt} 
    \renewcommand{\arraystretch}{1.2} 
    \caption{Performance comparison of fft-version of Llama-3b-instruct on different coefficient combinations for multiple metrics with clustering.}
    \vspace{-1em}
    \resizebox{\textwidth}{!}{
    \begin{tabular}{c@{\hspace{15pt}}ccccccccc}
        \toprule
        \multicolumn{3}{c}{\textbf{Hyperparameter}} & \multirow{2}{*}{\textbf{Train Loss}} & \multirow{2}{*}{\textbf{Eval. Loss}} & \multicolumn{2}{c}{\textbf{MT-Bench}} & \multicolumn{3}{c}{\textbf{Arena-Hard}} \\
        \cmidrule(lr){1-3} \cmidrule(r){6-7} \cmidrule(l){8-10} 
        Diff & Sep & Stab & & & \textbf{Score} & \textbf{Avg. Tokens} & \textbf{Score} & \textbf{95\% CI} & \textbf{Avg. Tokens} \\
        \midrule
        1 & 1 & 1   & 0.312 & 0.715 & 6.913  & 307  & 81.8  & $(-0.5, 0.8)$  & 266 \\
        1 & -1 & 1    & 0.368 & 0.803 & 6.625  & 292  & 84.2  & $(-0.7, 1.0)$  & 269 \\
        1 & 1 & 2   & 0.325 & 0.717 & \textbf{7.103}  & 328  & \textbf{85.5}  & $(-0.8, 1.1)$  & 271 \\
        1 & 1 & -1  & 0.294 & 0.617 & 6.650  & 298  & 82.7  & $(-1.5, 1.4)$  & 278 \\
        1 & 1 & 1.5 & 0.338 & 0.721 & 6.850  & 312  & 84.7  & $(-1.6, 1.3)$  & 266 \\
        1 & -1 & 1.5  & 0.391 & 0.795 & 6.781  & 286  & 83.0  & $(-1.4, 1.4)$  & 270 \\
        -1 & -1 & 1     & 0.354 & 0.707 & 6.781  & 308  & 81.9  & $(-1.5, 1.3)$  & 275 \\
        -1 & -1 & 2     & 0.355 & 0.742 & 6.838  & 297  & 84.8  & $(-1.3, 1.2)$  & 275 \\
        -1 & -1 & 1.5   & 0.351 & 0.754 & 6.638  & 289  & 81.8  & $(-1.3, 1.3)$  & 276 \\
        \bottomrule
    \end{tabular}
    }
    \label{fft compare - our metrics + cluster + multimetric - llama 3b instruct - swp}
\end{table*}

\begin{table*}[h]
\centering
    \small
    \setlength{\tabcolsep}{10pt}  
    \renewcommand{\arraystretch}{1.2}   
    \caption{Performance comparison of fft-version of Qwen-3b-instruct with different coefficient combinations for multiple metrics.}
    \vspace{-1em}
    \resizebox{\textwidth}{!}{
    \begin{tabular}{c@{\hspace{15pt}}ccccccccc}
        \toprule[1.5pt]
        \multicolumn{3}{c}{\textbf{Hyperparameter}} & \multirow{2}{*}{\textbf{Train Loss}} & \multirow{2}{*}{\textbf{Eval. Loss}} & \multicolumn{2}{c}{\textbf{MT-Bench}} & \multicolumn{3}{c}{\textbf{Arena-Hard}} \\
        \cmidrule(lr){1-3} \cmidrule(r){6-7} \cmidrule(l){8-10}
        Diff & Sep & Stab & & & \textbf{Score} & \textbf{Avg. Tokens} & \textbf{Score} & \textbf{95\% CI} & \textbf{Avg. Tokens} \\
        \midrule
        1 & 1 & 1   & 0.354 & 0.776 & 6.856  & 359  & 83.6  & $(-1.7, 1.2)$  & 259 \\
        1 & -1 & 1    & 0.432 & 0.861 & 7.138  & 383  & 81.6  & $(-1.4, 1.5)$  & 259 \\
        1 & 1 & 2   & 0.371 & 0.776 & 7.131  & 366  & \textbf{85.2}  & $(-1.2, 1.1)$  & 262 \\
        1 & 1 & -1  & 0.310 & 0.645 & 7.231  & 376  & 82.3  & $(-1.6, 1.5)$  & 261 \\
        1 & 1 & 1.5 & 0.369 & 0.755 & 6.981  & 387  & 83.6  & $(-2.0, 1.2)$  & 260 \\
        1 & -1 & 1.5  & 0.430 & 0.872 & \textbf{7.371}  & 390  & 82.4  & $(-1.7, 1.5)$  & 260 \\
        -1 & -1 & 1     & 0.431 & 0.874 & 7.025  & 397  & 81.9  & $(-1.1, 1.9)$  & 260 \\
        -1 & -1 & 2     & 0.431 & 0.888 & 6.963  & 377  & 80.6  & $(-1.8, 1.5)$  & 259 \\
        -1 & -1 & 1.5   & 0.433 & 0.869 & 6.956  & 377  & 82.4  & $(-1.8, 1.3)$  & 260 \\
        \bottomrule[1.5pt]
    \end{tabular}
    }
    \label{fft compare - our metrics + cluster + multimetric - qwen 2.5 instruct}
\end{table*}

\begin{table*}[h]
\centering
    \small
    \setlength{\tabcolsep}{10pt}  
    \renewcommand{\arraystretch}{1.2}   
    \caption{Performance comparison of fft-version of Llama-3b with different coefficient combinations for multiple metrics.}
    \vspace{-1em}
    \resizebox{\textwidth}{!}{
    \begin{tabular}{c@{\hspace{15pt}}ccccccccc}
        \toprule[1.5pt]
        \multicolumn{3}{c}{\textbf{Hyperparameter}} & \multirow{2}{*}{\textbf{Train Loss}} & \multirow{2}{*}{\textbf{Eval. Loss}} & \multicolumn{2}{c}{\textbf{MT-Bench}} & \multicolumn{3}{c}{\textbf{Arena-Hard}} \\
        \cmidrule(lr){1-3} \cmidrule(r){6-7} \cmidrule(l){8-10}
        Diff & Sep & Stab & & & \textbf{Score} & \textbf{Avg. Tokens} & \textbf{Score} & \textbf{95\% CI} & \textbf{Avg. Tokens} \\
        \midrule
        1 & 1 & 1   & 0.437 & 0.901 & 4.800  & 306  & 80.8  & $(-1.3, 1.6)$  & 289 \\
        1 & -1 & 1    & 0.497 & 1.007 & 5.019  & 319  & 80.3  & $(-2.2, 2.1)$  & 290 \\
        1 & 1 & 2   & 0.454 & 0.904 & 4.613  & 282  & 82.1  & $(-1.8, 1.8)$  & 290 \\
        1 & 1 & -1  & 0.416 & 0.786 & 4.669  & 283  & \textbf{83.0}  & $(-1.6, 2.0)$  & 289 \\
        1 & 1 & 1.5 & 0.449 & 0.908 & 4.731  & 276  & 75.7  & $(-1.9, 2.4)$  & 290 \\
        1 & -1 & 1.5  & 0.496 & 1.016 & \textbf{5.125}  & 309  & 80.6  & $(-2.4, 1.6)$  & 290 \\
        -1 & -1 & 1     & 0.469 & 0.973 & 5.050  & 307  & 80.7  & $(-1.8, 1.2)$  & 289 \\
        -1 & -1 & 2     & 0.469 & 0.968 & 4.719  & 268  & 81.6  & $(-1.2, 1.1)$  & 290 \\
        -1 & -1 & 1.5   & 0.469 & 0.968 & 4.588  & 291  & 80.0  & $(-2.0, 1.8)$  & 290 \\
        \bottomrule[1.5pt]
    \end{tabular}
    }
    \label{fft compare - our metrics + cluster + multimetric - llama3.2 base}
\end{table*}

\begin{table*}[h]
\centering
    \small
    \setlength{\tabcolsep}{10pt}
    \renewcommand{\arraystretch}{1.2}   
    \caption{Performance comparison of fft-version of Qwen-3b with different coefficient combinations for multiple metrics.}
    \vspace{-1em}
    \resizebox{\textwidth}{!}{
    \begin{tabular}{c@{\hspace{15pt}}ccccccccc}
        \toprule[1.5pt]
        \multicolumn{3}{c}{\textbf{Hyperparameter}} & \multirow{2}{*}{\textbf{Train Loss}} & \multirow{2}{*}{\textbf{Eval. Loss}} & \multicolumn{2}{c}{\textbf{MT-Bench}} & \multicolumn{3}{c}{\textbf{Arena-Hard}} \\
        \cmidrule(lr){1-3} \cmidrule(r){6-7} \cmidrule(l){8-10}
        Diff & Sep & Stab & & & \textbf{Score} & \textbf{Avg. Tokens} & \textbf{Score} & \textbf{95\% CI} & \textbf{Avg. Tokens} \\
        \midrule
        1 & 1 & 1   & 0.335 & 0.820 & 5.806  & 354  & 77.8  & $(-0.9, 1.8)$  & 249 \\
        1 & -1 & 1    & 0.399 & 0.917 & 6.544  & 415  & 78.0  & $(-1.7, 1.6)$  & 249 \\
        1 & 1 & 2   & 0.347 & 0.823 & 6.288  & 383  & \textbf{79.9}  & $(-1.6, 1.8)$  & 252 \\
        1 & 1 & -1  & 0.300 & 0.686 & 6.175  & 386  & 77.7  & $(-1.6, 2.4)$  & 253 \\
        1 & 1 & 1.5 & 0.343 & 0.804 & 5.981  & 348  & 77.5  & $(-1.6, 1.4)$  & 246 \\
        1 & -1 & 1.5  & 0.397 & 0.931 & \textbf{6.625}  & 309  & 78.0  & $(-1.6, 2.0)$  & 290 \\
        -1 & -1 & 1     & 0.397 & 0.916 & 6.188  & 410  & 79.2  & $(-1.5, 1.8)$  & 249 \\
        -1 & -1 & 2     & 0.397 & 0.923 & 6.331  & 391  & 78.8  & $(-1.3, 1.7)$  & 248 \\
        -1 & -1 & 1.5   & 0.397 & 0.927 & 6.325  & 380  & 77.7  & $(-1.9, 1.9)$  & 252 \\
        \bottomrule[1.5pt]
    \end{tabular}
    }
    \label{fft compare - our metrics + cluster + multimetric - qwen2.5 base}
\end{table*}

\begin{table*}[htbp]
\centering
\setlength{\tabcolsep}{4pt}  
\caption{Performance comparison of Llama-3b-instruct models with different fine-tuning methods}
\vspace{-1em}
\begin{tabular}{ccccccccc}
\toprule[1.5pt]
\multirow{2}{*}{\textbf{Benchmark}} 
& \multirow{2}{*}{\textbf{Random}} 
& \multicolumn{2}{c}{\textbf{Difficulty}} 
& \multicolumn{2}{c}{\textbf{Separability}} 
& \multicolumn{2}{c}{\textbf{Stability}} \\
& & \textbf{$\downarrow$} & \textbf{$\uparrow$} 
& \textbf{$\downarrow$} & \textbf{$\uparrow$} 
& \textbf{$\downarrow$} & \textbf{$\uparrow$} \\
\midrule
\multicolumn{8}{c}{\textbf{SFT}} \\
\midrule
MT-Bench & 6.200 & 6.388 & \textbf{6.648} & 5.937 & 6.581 & 6.225 & 6.625 \\
Arena-Hard & 74.4 & 76.5 & \textbf{80.5} & 77.9 & 80.0 & 75.8 & 77.4 \\
Arena-Hard-95\%CI & (-1.0, 1.5) & (-1.6, 1.5) & (-0.9, 1.3) & (-1.5, 1.7) & (-1.3, 1.2) & (-1.3, 0.9) & (-1.5, 1.1) \\
\midrule
\multicolumn{8}{c}{\textbf{DPO}} \\
\midrule
MT-Bench  & 6.463 & 6.431 & 6.768 & 6.431 & 6.418 & 6.256 & \textbf{6.818} \\
Arena-Hard  & 74.2 & 75.1 & 77.3 & 76.1 & \textbf{78.5} & 73.2 & 76.2 \\
Arena-Hard-95\%CI  & (-1.8, 1.6) & (-1.6, 1.6) & (-1.6, 1.7) & (-1.9, 1.9) & (-1.5, 1.4) & (-1.4, 1.3) & (-1.9, 1.5) \\
\midrule
\multicolumn{8}{c}{\textbf{SimPO}} \\
\midrule
MT-Bench  & 6.950 & 6.425 & \textbf{7.137} & 6.518 & 7.043 & 6.675 & 6.931 \\
Arena-Hard  & 78.7 & 78.0 & 78.8 & 78.2 & \textbf{79.7} & 76.0 & 75.5 \\
Arena-Hard-95\%CI  & (-2.5, 2.0) & (-2.5, 3.1) & (-0.9, 1.2) & (-1.6, 0.8) & (-5.4, 6.5) & (-1.3, 1.1) & (-5.7, 6.2) \\
\midrule
\multicolumn{8}{c}{\textbf{ORPO}} \\
\midrule
MT-Bench  & 6.412 & 6.450 & 6.450 & \textbf{6.525} & 6.431 & 6.312 & 6.400 \\
Arena-Hard  & 73.7 & 73.2 & 73.7 & 73.3 & 74.6 & 73.2 & \textbf{75.6} \\
Arena-Hard-95\%CI  & (-2.1, 2.2) & (-2.2, 1.8) & (-1.5, 2.0) & (-1.9, 1.8) & (-2.0, 2.2) & (-2.1, 2.2) & (-1.8, 2.2) \\
\bottomrule[1.5pt]
\end{tabular}
\label{training method comparison - llama 3b instruct - swp}
\end{table*}

\begin{table*}[htbp]
\centering
\setlength{\tabcolsep}{4pt}  
\caption{Performance comparison of lora-version of Llama-3b-instruct models with different reward-models}
\vspace{-1em}
\resizebox{\textwidth}{!}{
\begin{tabular}{cccccccccc}
\toprule[1.5pt]
\multirow{2}{*}{\textbf{Benchmark}} 
& \multicolumn{2}{c}{\textbf{Difficulty}} 
& \multicolumn{2}{c}{\textbf{Separability}} 
& \multicolumn{2}{c}{\textbf{Stability}}
& \multicolumn{2}{c}{\textbf{Reward-Score}} \\
& \textbf{$\downarrow$} & \textbf{$\uparrow$} 
& \textbf{$\downarrow$} & \textbf{$\uparrow$} 
& \textbf{$\downarrow$} & \textbf{$\uparrow$} 
& \textbf{$\downarrow$} & \textbf{$\uparrow$} \\
\midrule
\multicolumn{9}{c}{\textbf{ArmoRM-Llama3-8B-v0.1}} \\
\midrule
MT-Bench & 6.625 & \textbf{6.687} & 6.468 & 6.493 & 6.375 & 6.431 & 4.037 & 6.512\\
Arena-Hard & \textbf{81.7} & 78.6 & 74.3 & 75.6 & 77.3 & 80.0 & 57.8 & 83.2 \\
Arena-Hard-95\%CI  & (-2.0, 1.8) & (-1.8, 1.8) & (-1.8, 2.1) & (-2.0, 1.6) & (-1.8, 2.0) & (-1.0, 1.8) & (-2.0, 1.9) & (-1.5, 1.9)\\
\midrule
\multicolumn{9}{c}{\textbf{Skywork-Reward-Llama-3.1-8B}} \\
\midrule
MT-Bench  & 6.456 & 6.688 & 6.100 & 6.725 & 6.131 & \textbf{6.866} & 4.012 & 6.675\\
Arena-Hard  & 69.6 & 76.8 & 69.4 & 72.9 & 69.8 & 74.6 & 52.6 & \textbf{77.4} \\
Arena-Hard-95\%CI  & (-1.5,1.9) & (-1.8,1.4) & (-2.5,1.2) & (-1.6,1.5) & (-1.7,1.7) & (-1.7,2.0) & (-2.4, 2.0) & (-1.8, 2.1)\\
\midrule
\multicolumn{9}{c}{\textbf{Skywork-Reward-Gemma-2-27B}} \\
\midrule
MT-Bench  & 6.512 & 6.593 & 6.756 & 6.881 & 6.637 & 6.756 & 3.793 & \textbf{6.943} \\
Arena-Hard  & 76.2 & 78.2 & 75.4 & 80.2 & 79.7 & \textbf{83.6} & 56.1 & 79.6 \\
Arena-Hard-95\%CI  & (-1.6, 2.0) & (-1.6, 1.5) & (-2.1, 2.1) & (-1.7, 2.4) & (-1.4, 1.4) & (-1.9, 2.0) & (-2.1, 2.1) & (-1.6, 1.7) \\
\bottomrule[1.5pt]
\end{tabular}
}
\label{lora - llama 3b instruct - reward model ablation - swp}
\end{table*}

\begin{table*}[h]
    \centering
    \caption{Top-5 instruction-response pairs by \textit{Difficulty}.}
    \vspace{-1em}
    \label{tab:instruction_response_difficulty}
    \renewcommand{\arraystretch}{1.2}  
    \footnotesize
    \rowcolors{2}{white}{gray!20}  
    \begin{tabularx}{\textwidth}{X} 
        \toprule[1.5pt]
        \textbf{1. What is the name of the singer-songwriter known for his distinctive, raspy, belt-it-out-from-the-diaphragm vocals and soulful delivery, known for hits like "Swallowed in the Sea", " Wolfgang Amadeus!", and "Skinyalong"? Is it (a) Thom Yorke?, (b) Brandon Boyd?, (c) Dave Matthews? (d) Frank Ocean?} \\
        I think I have the answer: (a) Thom Yorke! Thom Yorke is the lead vocalis and principal songwriter of the British rock band Radiohead, known for his distinctive, emotive, and often raspy vocals \ldots \\
        \midrule
        \textbf{2. What is the percentage of hydrocarbons in the wastewater sludges generated by municipal wastewater treatment plants?} \\
        The percentage of hydrocarbons in wastewater sludges can vary widely depending on several factors, such as the type of wastewater treatment process used, industrial activities \ldots \\
        \midrule
        \textbf{3. What is a common typo that can result in innocuous words being misinterpreted as explicit content?} \\
        One common typo that can lead to misinterpretation is the accidental omission or substitution of a single letter or character. For example, "ass" instead of "as", "cock" instead of "cook" \ldots \\
        \midrule
        \textbf{4. What is the minimum cost of \( x \) that satisfies the equation \(\frac{1}{x+3}=3y\), with \(x, y\) positive integers?} \\
        We solve for \( x \) in terms of \( y \): \( x+3=\frac{1}{3y} \Rightarrow x=\frac{1}{3y}-3 \). Since \( x \) is a positive integer, the smallest such \( y \) is \( y=1 \), which gives \( x=2 \) \ldots \\
        \midrule
        \textbf{5. I have an integer \( x \) such that \( x^2 + 3x -2 = (x+1)(x+2) \). Simplify and find the value of \( x \).} \\
        Expanding the right side: \( (x+1)(x+2) = x^2 + 3x + 2 \). Setting equal to the left-hand side and simplifying: \( x^2 + 3x - 2 = x^2 + 3x + 2 \). Subtracting \ldots \\
        \bottomrule[1.5pt]
    \end{tabularx}
\end{table*}

\begin{table*}[h]
    \centering
    \caption{Top-5 instruction-response pairs by \textit{Separability}.}
    \vspace{-1em}
    \label{tab:instruction_response_seperability}
    \renewcommand{\arraystretch}{1.2}  
    \footnotesize
    \rowcolors{2}{white}{gray!20}  
    \begin{tabularx}{\textwidth}{X} 
        \toprule[1.5pt]
        \textbf{1. I am using Ubuntu 20.04 and I'm trying to install an older version of OpenCV (OpenCV 3.4.6) alongside my existing OpenCV 4.5.5 installation.} \\
        Installing an older version of OpenCV alongside a newer version requires careful handling to avoid conflicts. Here's a step-by-step guide to help you install OpenCV 3.4.6 alongside OpenCV 4.5.5 on Ubuntu 20.04: \ldots \\
        \midrule
        \textbf{2. You're a chef, and you're known for your exotic spices and unique flavor combinations. Can you create a dish that will wow a group of food critics who are known for being particularly discerning?} \\
        What an exciting challenge! I relish the opportunity to push the boundaries of culinary innovation and delight the palates of even the most discerning food critics. After some careful consideration, I propose the following dish: \ldots \\
        \midrule
        \textbf{3. I'm planning a road trip across the United States, and I'm looking for some fun and unique experiences to add to my itinerary. Can you suggest some unusual and offbeat attractions that I won't want to miss?} \\
        What an adventure! I'd be delighted to help you discover some quirky and offbeat attractions across the United States. Here are some unusual experiences to add to your road trip itinerary: \ldots \\
        \midrule
        \textbf{4. I'm planning a trip to Italy and want to be fitted for clothes at a high-end tailor in Italy. Does anyone know of any high-end tailors in Italy that offer made-to-measure or bespoke services?} \\
        Italy is renowned for its exceptional tailoring, and you'll find numerous high-end tailors offering made-to-measure and bespoke services. Here are some of the most esteemed tailors in Italy: \ldots \\
        \midrule
        \textbf{5. How to compile and install OpenCV 4.5.5 on Ubuntu 20.04?} \\
        Compiling and Installing OpenCV 4.5.5 on Ubuntu 20.04 Here's a comprehensive guide on compiling and installing OpenCV 4.5.5 from source on Ubuntu 20.04. **Prerequisites:** * **Ubuntu 20.04 LTS:** This guide \ldots \\
        \bottomrule[1.5pt]
    \end{tabularx}
\end{table*}

\begin{table*}[h]
    \centering
    \caption{Top-5 instruction-response pairs by \textit{Stability}.}
    \vspace{-1em}
    \label{tab:instruction_response_stability}
    \renewcommand{\arraystretch}{1.2}  
    \footnotesize
    \rowcolors{2}{white}{gray!20}  
    \begin{tabularx}{\textwidth}{X} 
        \toprule[1.5pt]
        \textbf{1. I am looking for a compiler (specifically, an assembler) for a historic computer, specifically an Interdata 8/32.} \\
        You're on a fascinating quest! Finding an assembler for a historic computer like the Interdata 8/32 can be quite a challenge. Here's a breakdown of where to look and what to consider: \ldots \\
        \midrule
        \textbf{2. I have a PHP function that currently grabs all videos from a vimeo video page. This function uses cURL to fetch the page content, then preg\_matches it for a specific class name to extract the video IDs.} \\
        It sounds like you're on the right track! Using cURL to fetch the page content and then using regular expressions (preg\_match) to extract video IDs is a common approach.   However, relying solely on HTML structure and class names can be fragile. \ldots \\
        \midrule
        \textbf{3. I'm a mobile developer and I have been experiencing an unusual issue in my android application where a RecyclerView adapter is not calling its onBindVieverHolder method when the RecyclerView is being scrolled more than 50-60 items high. How can I troubleshoot this issue?} \\
        Let's tackle this RecyclerView issue. Here's a breakdown of common causes and debugging strategies for when `onBindViewHolder` isn't firing as expected: **Understanding the Problem** The `onBindViewHolder` method is crucial for populating your RecyclerView items with data. If it's not being called beyond a certain scroll point,\ldots \\
        \midrule
        \textbf{4. What type of software exists for debugging and simulating network simulations in Python?} \\
        Python offers a rich ecosystem of tools for debugging and simulating network simulations. Here's a breakdown of some popular options, categorized by their focus: \ldots \\
        \midrule
        \textbf{5. I am using LinearGradientBrush class to draw a gradient in WPF, but the start and end points of the gradient are not showing correctly. The `StartPoint` property is set to `PointF(0, 0)` and the `EndPoint` property is set to `PointF(1, 1)`, but the gradient appears to be drawn all the way to the bottom-right corner of the control.} \\
        The issue you're facing is likely due to the way WPF's coordinate system works in conjunction with the `LinearGradientBrush`. **Understanding WPF's Coordinate System** WPF uses a coordinate system where (0, 0) is the top-left corner of the element, and (1, 1) is the bottom-right corner. This means your current settings are telling the gradient \ldots \\
        \bottomrule[1.5pt]
    \end{tabularx}
\end{table*}

\begin{table*}[h]
    \centering
    \captionsetup{justification=centering} 
    \caption{Top-5 instruction-response pairs by \method.}
    \vspace{-1em}
    \label{tab:instruction_response_crowd}
    \renewcommand{\arraystretch}{1.2}  
    \footnotesize
    \rowcolors{2}{white}{gray!20}  
    \resizebox{\textwidth}{!}{
    \begin{tabular}{p{\textwidth}} 
        \toprule[1.5pt]
        \textbf{1. What is GD-ROM and what is its importance in the history of video game consoles} \\
        GD-ROM (Gigabyte Disc Read-Only Memory) is a proprietary optical disc format developed by Yamaha for the Sega Dreamcast console in the late 1990s. It's a crucial part of the Dreamcast's architecture and played a significant role in the history of video game consoles.  Here's what makes GD-ROM important: \ldots \\
        \midrule
        \textbf{2. I am looking for a list of all the Oz books, including the original 14 that L. Frank Baum wrote, as well as the many spin-off and unofficial books that have been written.} \\
        You're in for a wild ride through Oz! It's impossible to list *every* Oz book ever written, as there are hundreds, but here's a breakdown to get you started: **1. The Original 14 by L. Frank Baum: \ldots \\
        \midrule
        \textbf{3. I want to install Kaldi on a Rust-based Debian Buster machine.} \\
        Let's get Kaldi up and running on your Rust-based Debian Buster system. Here's a comprehensive guide: **1. Prerequisites** \ldots \\
        \midrule
        \textbf{4. What are some of the lesser-known ski resorts in the US that are considered to be difficult to access?} \\
        You're looking for hidden gems, the kind of ski resorts that require a bit of effort to reach but reward you with uncrowded slopes and a sense of adventure! Here are a few lesser-known, hard-to-access ski resorts in the US: \ldots \\
        \midrule
        \textbf{5. What are some notable demolished or lost buildings and structures in NYC landmarks that are no longer standing?} \\
        What a great question! New York City's ever-changing landscape has led to the loss of many iconic buildings and structures over the years. Here are some notable demolished or lost NYC landmarks that are no longer standing: \ldots \\
        \bottomrule[1.5pt]
    \end{tabular}
    }
\end{table*}

\begin{figure*}
    \centering
    \begin{tcolorbox}[colframe=gray!50, colback=white, coltitle=black, title=\textbf{Example of Selected Instruction and Correspond Answer}, sharp corners=southwest, boxrule=0.5mm, width=0.8\textwidth, left=5mm, top=3mm, bottom=3mm]
    
    \textbf{Problem:}\\ What are 5 different methods to generate electricity. Not including hydroelectric, steam, geothermal, nuclear, or biomass. The method must not use any form of rotating generator where a coil is spun around magnets or the other way around. Turbines can not be used. No wind or tidal either.

    \vspace{5mm} 

    \textbf{Response:} \\4 main types and one alternative:

    **1. Photovoltaic (Solar) Cells:** These convert sunlight directly into electrical energy using semiconducting materials.

    * \textbf{How it works}: Solar cells contain photovoltaic material that excites electrons when exposed to light, causing them to flow through an external circuit creating electric current.

    **2. Fuel Cell Systems:** They produce electricity from chemical reactions between fuels like hydrogen gas \& oxygen.

    * \textbf{How they work}: Fuel cell systems split water molecules (H\(_2\)O) into protons, electrons \& hydrogen. The electrochemical reaction converts these components back together producing only heat as a waste product.

    **3. Silicone-Based Piezoelectric Generators:** These devices harness mechanical stress in silicone elastomers.

    * \textbf{How they function}: When subjected to pressure changes within their structure, piezoelectric properties cause separation of positive-negative charges leading to generation of voltage across internal electrodes.

    **4. Ceramic Capacitor Discharge Devices:** They utilize stored electrical energy in ceramic capacitors.

    * \textbf{How they operate}: A capacitor stores charge during the charging phase and then releases this accumulated potential difference rapidly upon discharge, triggering brief but intense bursts of high-voltage DC power.

    **Alternative Method - Photo Electro Chemical Cells:**

    This technology combines photo voltaics.
    
    \end{tcolorbox}
    \caption{Example of our selected instruction and correspond answer.}
    \label{fig: example}
\end{figure*}

\begin{figure*}[h]
    \centering
     
    \begin{subfigure}{0.32\textwidth}
        \captionsetup{position=above, aboveskip=0pt, belowskip=0pt} 
        \caption*{0.25$k$}
        \includegraphics[width=\linewidth]{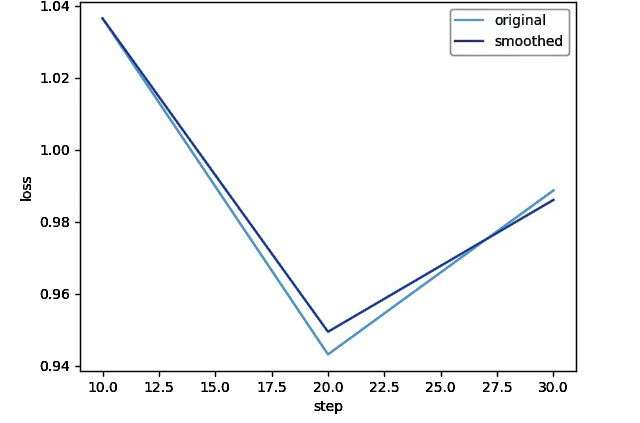}
    \end{subfigure}\hfill
    \begin{subfigure}{0.32\textwidth}
        \captionsetup{position=above, aboveskip=0pt, belowskip=0pt}
        \caption*{0.5$k$}
        \includegraphics[width=\linewidth]{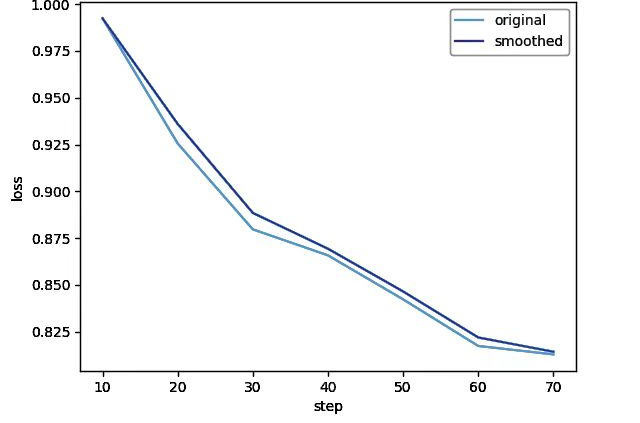}
    \end{subfigure}\hfill
    \begin{subfigure}{0.32\textwidth}
        \captionsetup{position=above, aboveskip=0pt, belowskip=0pt}
        \caption*{1$k$}
        \includegraphics[width=\linewidth]{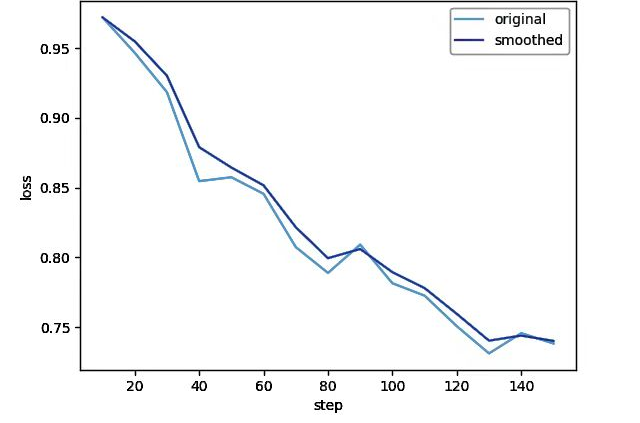}
    \end{subfigure}

    \begin{subfigure}{0.32\textwidth}
        \captionsetup{position=above, aboveskip=0pt, belowskip=0pt}
        \caption*{2$k$}
        \includegraphics[width=\linewidth]{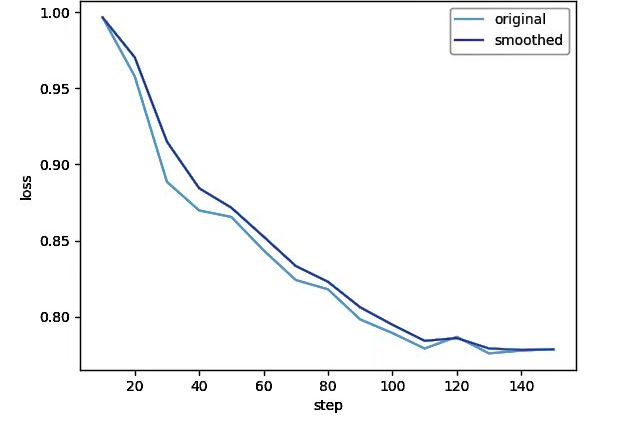}
    \end{subfigure}\hfill
    \begin{subfigure}{0.32\textwidth}
        \captionsetup{position=above, aboveskip=0pt, belowskip=0pt}
        \caption*{3$k$}
        \includegraphics[width=\linewidth]{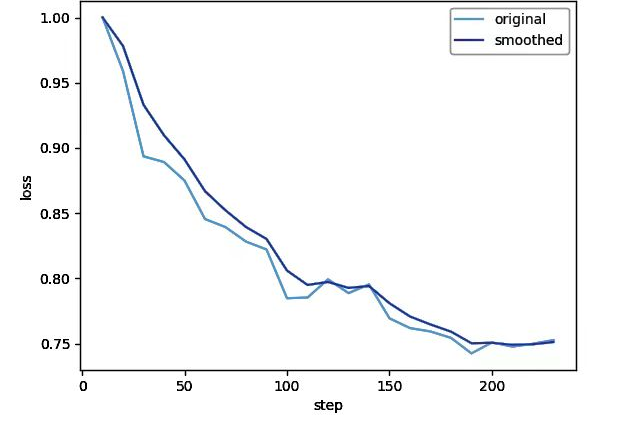}
    \end{subfigure}\hfill
    \begin{subfigure}{0.32\textwidth}
        \captionsetup{position=above, aboveskip=0pt, belowskip=0pt}
        \caption*{4$k$}
        \includegraphics[width=\linewidth]{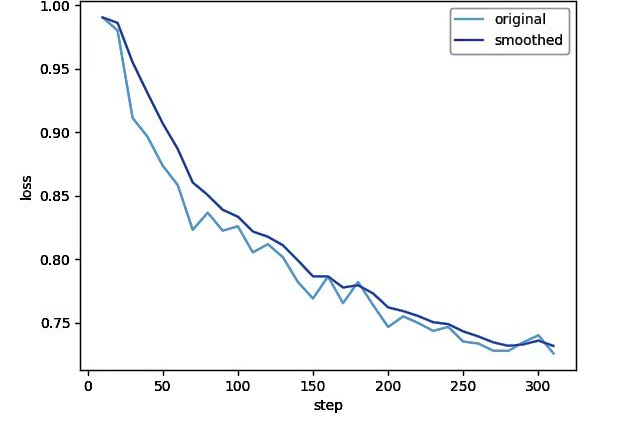}
    \end{subfigure}

    \begin{subfigure}{0.32\textwidth}
        \captionsetup{position=above, aboveskip=0pt, belowskip=0pt}
        \caption*{5$k$}
        \includegraphics[width=\linewidth]{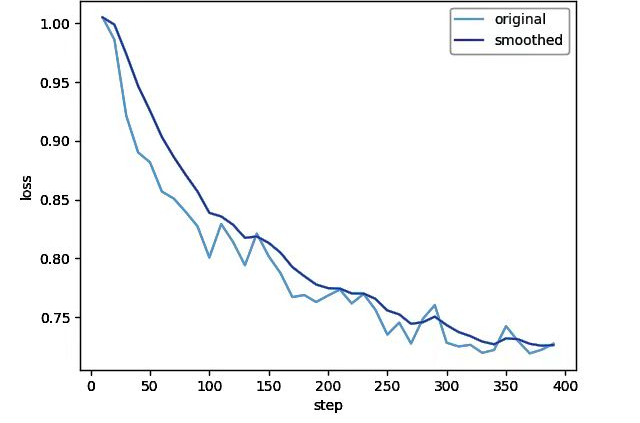}
    \end{subfigure}\hfill
    \begin{subfigure}{0.32\textwidth}
        \captionsetup{position=above, aboveskip=0pt, belowskip=0pt}
        \caption*{6$k$}
        \includegraphics[width=\linewidth]{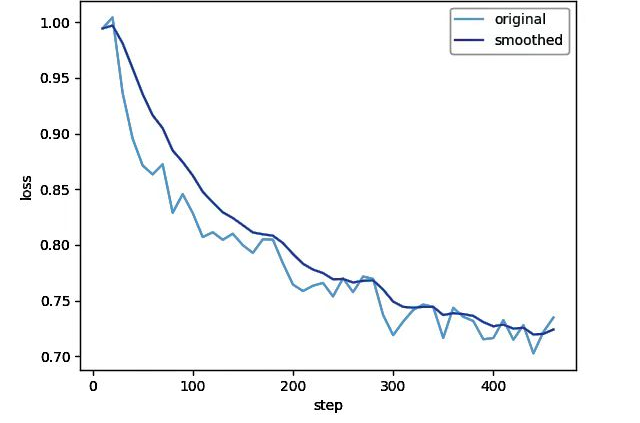}
    \end{subfigure}\hfill
    \begin{subfigure}{0.32\textwidth}
        \captionsetup{position=above, aboveskip=0pt, belowskip=0pt}
        \caption*{7$k$}
        \includegraphics[width=\linewidth]{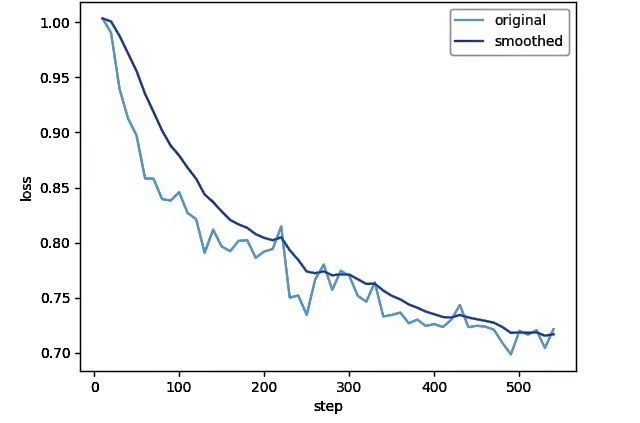}
    \end{subfigure}

    \begin{subfigure}{0.32\textwidth}
        \captionsetup{position=above, aboveskip=0pt, belowskip=0pt}
        \caption*{8$k$}
        \includegraphics[width=\linewidth]{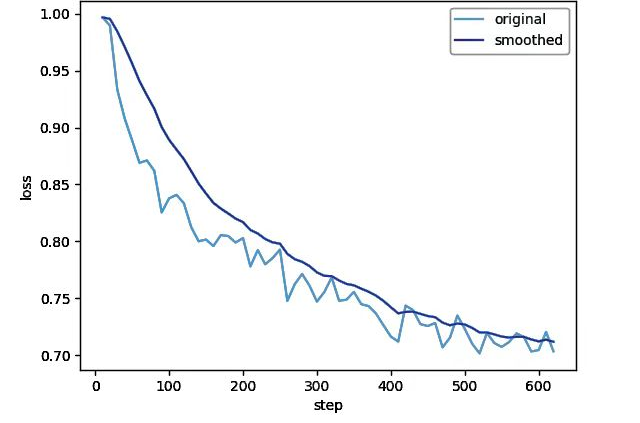}
    \end{subfigure}\hfill
    \begin{subfigure}{0.32\textwidth}
        \captionsetup{position=above, aboveskip=0pt, belowskip=0pt}
        \caption*{9$k$}
        \includegraphics[width=\linewidth]{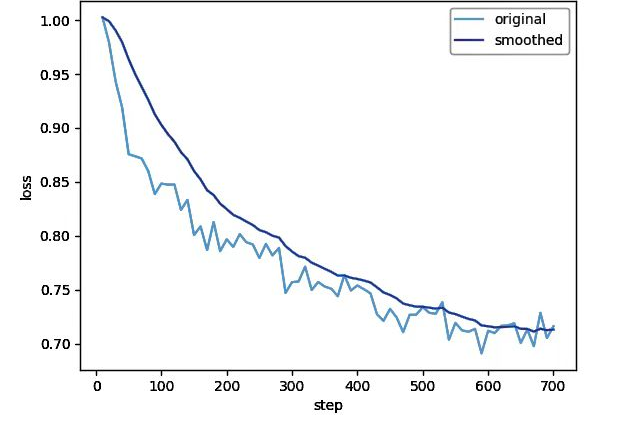}
    \end{subfigure}\hfill
    \begin{subfigure}{0.32\textwidth}
        \captionsetup{position=above, aboveskip=0pt, belowskip=0pt}
        \caption*{10$k$}
        \includegraphics[width=\linewidth]{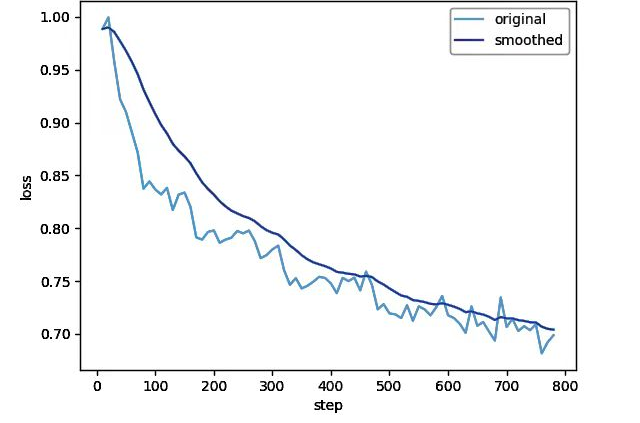}
    \end{subfigure}

    \caption{Lora train loss of training Llama-3b by using different sizes of randomly chosen data.}
    \label{fig:lora_train_loss}
\end{figure*}

\end{document}